\documentclass[a4paper,twoside]{article}

\usepackage{times}
\usepackage[outdir=./]{epstopdf}
\usepackage{epsfig}
\usepackage{graphicx}
\usepackage{amsmath}
\usepackage{amssymb}

\usepackage{comment}
\usepackage{wrapfig}
\usepackage{booktabs, multirow}        
\usepackage[dvipsnames]{xcolor}  
\usepackage{xfrac}
\usepackage{caption,subcaption}

\usepackage{calc}
\usepackage{amstext}
\usepackage{amsthm}
\usepackage{multicol}
\usepackage{pslatex}

\usepackage{apalike}
\usepackage{SCITEPRESS}     

\newcommand{\DZ}[1]{\textcolor{ForestGreen}{DZ: #1}}

\begin{document}

\title{Scan2Part: Fine-grained and Hierarchical Part-level~Understanding~of~Real-World 3D~Scans}

\author{\authorname{Alexandr Notchenko,Vladislav Ishimtsev, Alexey Artemov, Vadim Selyutin, Emil Bogomolov, Evgeny Burnaev}
\affiliation{Skolkovo Institute of Science and Technology, Moscow, Russian Federation}
\email{\{a.notchenko,vladislav.ishimtsev, a.artemov, vadim.selyutin, E.Bogomolov, E.Burnaev\}@skoltech.ru}
}






\abstract{
We propose Scan2Part, a method to segment individual parts of objects in real-world, noisy indoor RGB-D scans.
To this end, we vary the part hierarchies of objects in indoor scenes and explore their effect on scene understanding models.
Specifically, we use a sparse U-Net-based architecture that captures the fine-scale detail of the underlying 3D scan geometry by leveraging a multi-scale feature hierarchy.
In order to train our method, we introduce the Scan2Part dataset, which is the first large-scale collection providing detailed semantic labels at the part level in the real-world setting.
In total, we provide 242,081 correspondences between 53,618 PartNet parts of 2,477 ShapeNet objects and 1,506 ScanNet scenes, at two spatial resolutions of 2\,cm$^3$ and 5\,cm$^3$.
%
As output, we are able to predict fine-grained per-object part labels, even when the geometry is coarse or partially missing.
%
%
Overall, we believe that both our method as well as newly introduced dataset is a stepping stone forward towards  structural understanding of real-world 3D environments.
}

\onecolumn \maketitle

\section{Introduction}

In the recent years, a wide variety of consumer-grade RGB-D cameras such as Intel Real Sense~\cite{keselman2017intel}, Microsoft Kinect~\cite{zhang2012microsoft}, or smartphones equipped with depth sensors, enabled inexpensive and rapid RGB-D data acquisition. 
Increasing availability of large, labeled datasets (e.g.,~\cite{chang2017matterport3d,dai2017scannet}) made possible development of deep learning methods for 3D object classification and semantic segmentation.
At the same time, acquired 3D data is often incomplete and noisy; while one can identify and segment the objects in the scene, reconstructing high-quality geometry of objects remains a challenging problem.

An important class of approaches, e.g., recent work \cite{avetisyan2019scan2cad}, uses a large dataset of clean, labeled geometric shapes \cite{chang2015shapenet}, for classification/segmentation associating the input point or voxel data with object labels from the dataset, along with adapting geometry to 3D data.
This approach ensures that the output geometry has high quality, and is robust with respect to noise and missing data in the input.  
At the same time, a ``flat'' classification/segmentation approach, with each object in the database corresponding to a separate label and matched to a subpart of the input data corresponding to the whole object, does not scale well as the number of classes grows and often runs into difficulties in the cases of extreme occlusion (only a relatively small part of an object is visible). 
Significant improvements can be achieved by considering object \emph{parts}, or, more generally, part hierarchies. 
Part-based segmentation of 3D datasets promises to offer a significant improvement in a variety of tasks such as finding the best matching shape in the dataset or recognizing objects from highly incomplete data (e.g., from one or two parts).

Man-made environments (e.g., indoor scenes) commonly consist of objects that naturally form  \emph{part hierarchies} where objects and their parts can be divided into finer parts.
In our work, we use scene and object representation based on such part hierarchies and focus on the key problem of semantic part segmentation of separate objects in the scenes, enabling further improvements in dataset-based reconstruction. 
To this end, we construct a volumetric part-labeled dataset of scanned 3D data suitable for machine learning applications. 
We use this new dataset to explore the limits of part-based semantic segmentation by training a variety of sparse 3D convolutional neural networks (CNNs) in multiple setups.



%

Our contributions include: 
\begin{enumerate}
\item A new method Scan2Part, aiming to segment volumetric objects into semantic parts and their instances, leveraging a sparse volumetric 3D CNN model trained on a large-scale part-annotated collection of objects.

\item A new dataset Scan2Part, composed of 1,506 3D reconstructions of real-world scenes with 2,477 aligned 3D CAD models represented by real-world 3D geometry and annotated using hierarchical annotation, which links 3D scene reconstruction with part-annotation of indoor objects.

\end{enumerate}

\section{Related Work}
\label{sec:related}

\noindent \textbf{Deep learning for 3d scene understanding.}
\label{related:scene-understanding}
Deep learning has been applied for semantic 3D scene understanding in a variety of ways, and we only review the core work related to the semantic and instance segmentation of 3D scenes. 
Most relevant to our work, deep learning approaches have been used on 3D reconstructions of scenes represented by \emph{3D volumetric grids} ~\cite{dai2017scannet,dai20183dmv,hou20193d,liu2019masc}.
For volumetric grid, 3D convolutions may be defined analogously to 2D convolutions in image domains, giving rise to 3D convolutional neural nets (3D CNNs).
Memory requirements have been addressed by adaptive data structures~\cite{wang2017cnn}.
Similar to this body of work, we operate on volumetric representations of 3D scenes, but perform segmentation of individual parts. 

Methods operating on raw point clouds provide an alternative to volumetric 3D CNNs by constructing an appropriate operations directly on \emph{unstructured point clouds} for a variety of applications including semantic labeling (e.g.,~\cite{qi2017pointnet,qi2017pointnet++,klokov2017escape,wang2018sgpn,wang2019dynamic}). Most recently, instance~\cite{elich20193d,liang20193d,elich20193d,yi2019gspn,yang2019learning,zhang2019point,engelmann20203d} and joint semantic-instance~\cite{wang2019associatively,pham2019jsis3d} segmentation tasks on point clouds have been considered closely. 
While point-based methods require less computations, learning with irregular structures such as point clouds is challenging.
To segment instances, recently proposed volumetric and point-based approaches use metric learning to extract per-point embeddings that are subsequently grouped into object instances~\cite{elich20193d,yi2019gspn,lahoud20193d,liu2019masc}. 
We take advantage of this mechanism in our instance segmentation methods.


Part-aware segmentation methods commonly focus on meshes or complete, clean point clouds constructed from 3D CAD models. 
The most closely related work is semantic parts labeling (e.g.,~\cite{yi2016scalable,wang2017cnn,qi2017pointnet,mo2019partnet,yi2019gspn,zhang2019point}) and part instance segmentation~\cite{zhang2019point} for voxelized or point-sampled 3D shapes. Other works focus on leveraging parts structure of clean shapes for co-segmentation~\cite{chen2019bae,zhu2019cosegnet}, hierarchical mesh segmentation~\cite{yi2017learning}, shape assembly/generation~\cite{mo2019structurenet,wu2019sagnet,wu2019pq,mo2020pt2pc}, geometry abstraction~\cite{russell2009associative,li2017grass,sun2019learning}, and other applications.
In comparison, our focus is on learning part-based semantic and instance segmentation of noisy and fragmented real-world 3D scans. 
Very recently, initial approaches to semantic 3D segmentation have been proposed~\cite{bokhovkin2021towards,uy2019revisiting} but for a significantly less extensive part hierarchy. 
More specifically,~\cite{bokhovkin2021towards} targets predicting part hierarchy at object and coarse parts levels, discarding smaller parts altogether; in contrast, we are able to predict parts at finer levels in the hierarchy.

Other methods have been studied in the context of 3D scene segmentation, such as complementing CNNs with conditional random fields~\cite{pham2019jsis3d,pham2019real,wang2017cnn}, however, these are beyond the scope of this paper.

\begin{figure*}[!t]
\centering
\includegraphics[width=0.9\textwidth]{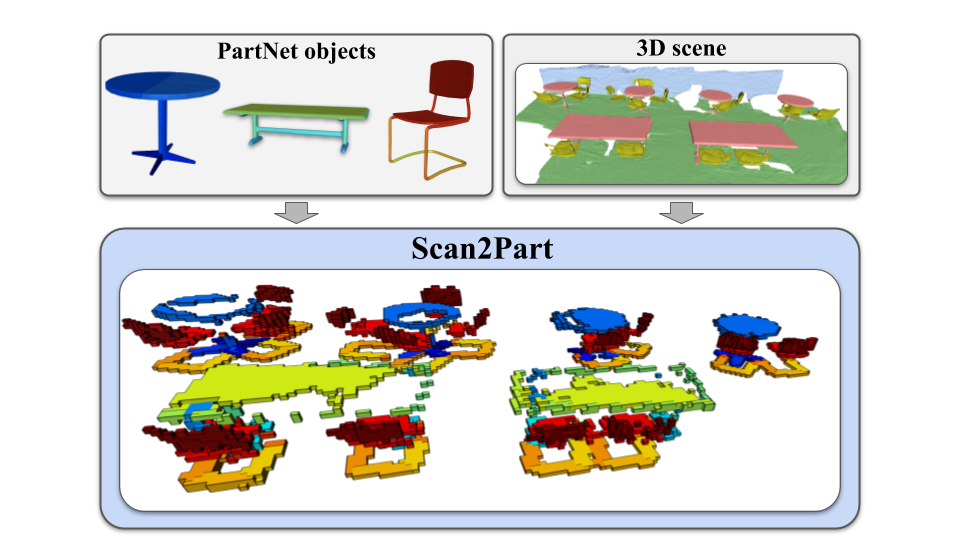}
\caption{A pipeline for obtaining Scan2Part dataset.
We project the PartNet \cite{mo2019partnet} labels to the ShapeNet \cite{chang2015shapenet} coordinate system (left), then use alignments in Scan2CAD \cite{avetisyan2019scan2cad} (bottom) to map labels to real indoor scenes from ScanNet \cite{dai2017scannet} (right).}
\label{fig:dataset_schema}
\end{figure*}

\noindent \textbf{3D scene understanding datasets.}
\label{related:datasets}
A body of work focuses on rendering-based methods, aiming to realistic 3D scenes procedurally~\cite{2012-scenesynth,handa2016understanding,song2016ssc,McCormac:etal:ICCV2017,InteriorNet18,garcia2018robotrix}.
Such datasets can in principle provide arbitrarily fine semantic labels but commonly suffer from the reality gap caused by synthetic images; in contrast, our proposed dataset is built by transferring part annotations to real-world noisy scans.
Recent advances in RGB-D sensor technology have resulted in the development of a variety of 3D datasets capturing real 3D scenes~\cite{armeni20163d,hua2016scenenn,dai2017scannet,chang2017matterport3d,2017arXiv170201105A,replica19arxiv}, however, none of these provide part-level object annotations.
In contrast, our dataset provides semantic and instance part labels for a large-scale collection of indoor 3D reconstructions.
ScanObjectNN~\cite{uy2019revisiting} provides parts annotation for objects in real-world scenes, however, it does not include annotations in occluded regions and only specifies parts labeling at the coarsest levels in the parts hierarchy.
Thus, this collection thus does not allow reasoning about the fine grained structure of the objects; differently, our data construction approach allows to flexibly select part representation levels, and our experiments include results for 13, 36, and 79\,part categories at first through third levels in the hierarchy.




Early collections of part-annotated meshes~\cite{Chen:2009:ABF} are limited by their relatively smaller scale.
With the introduction of a comprehensive ShapeNet benchmark~\cite{chang2015shapenet}, a coarse semantic part annotation has been created using active learning~\cite{yi2016scalable}. 
More recently, a large-scale effort to systematically annotate 3D shapes within a coherent hierarchy was presented~\cite{mo2019partnet}.
Still, none of these CAD-based collections include real-world 3D data, limiting their potential use. 
Our benchmark is designed to address this reality gap.

Large-scale 3D understanding datasets commonly require costly manual annotations by tens to hundreds of expert crowd workers (annotators), preceded by the development of custom labeling software~\cite{armeni20163d,hua2016scenenn,dai2017scannet,chang2017matterport3d,2017arXiv170201105A,replica19arxiv,yi2016scalable,mo2019partnet}.
Moreover, annotating parts in 3D objects from scratch is connected to inherent ambiguity in part definitions, as revealed by~\cite{yi2016scalable,mo2019partnet}. This challenge is even more pronounced for noisy, incomplete 3D scans produced by RGB-D fusion.
We have chosen to instead build our Scan2Part dataset fully automatically by leveraging correspondences between four publicly available 3D collections: ScanNet~\cite{dai2017scannet}, Scan2CAD~\cite{avetisyan2019scan2cad}, ShapeNet~\cite{chang2015shapenet}, and PartNet~\cite{mo2019partnet} datasets.
As a result, we (1) become free from ambiguity in part definitions by re-using consistent, well-defined labels from~\cite{mo2019partnet}, and (2) are able to compute appropriate levels of semantic detail for our benchmark without manual re-labeling.

\noindent \textbf{Assembly from parts and hierarchy.}
A lot of researchers over the last 20 years came to the idea that scenes and images are better represented as discrete structures with relational and hierarchical properties.
New datasets that make explicit relations on visual objects were created recently \cite{krishna2017visual}, spurring new research in scene graphs \cite{DBLP:journals/corr/abs-1804-01622,Xu_2017_CVPR} and reconstruction. 
In cognitive science, it have been conjectured for a long time that ability to compose complex objects and scenes from parts is a fundamental part of human perception \cite{hoffman1984parts,biederman1987recognition}. The concept of "Recognition-by-components" is closely related to "analysis-by-synthesis" \cite{yuille2006vision,yildirim2015efficient} approach in machine learning. 
Mumford and Zhu in \cite{zhu2006stochastic} assume that spatial scenes can be defined by a "grammar" of spatial objects and geometric primitives, similar to sentences in natural language, where a sequence of words can have multiple parsing trees providing multiple explanations to single sentence. Multiple solutions are expected because solving an under-constrained inverse problem (in this case inverse graphics problem) can have multiple solutions. The solution can be made unique either with stronger priors or can be resolved in downstream processing if uncertainty is propagated.



\section{Method}
\label{sec:method}

\subsection{Dataset Construction}
\label{dataset:construction}

\begin{figure*}[t]
\label{fig:dataset_teaser}
\centering
\includegraphics[width=0.99\textwidth]{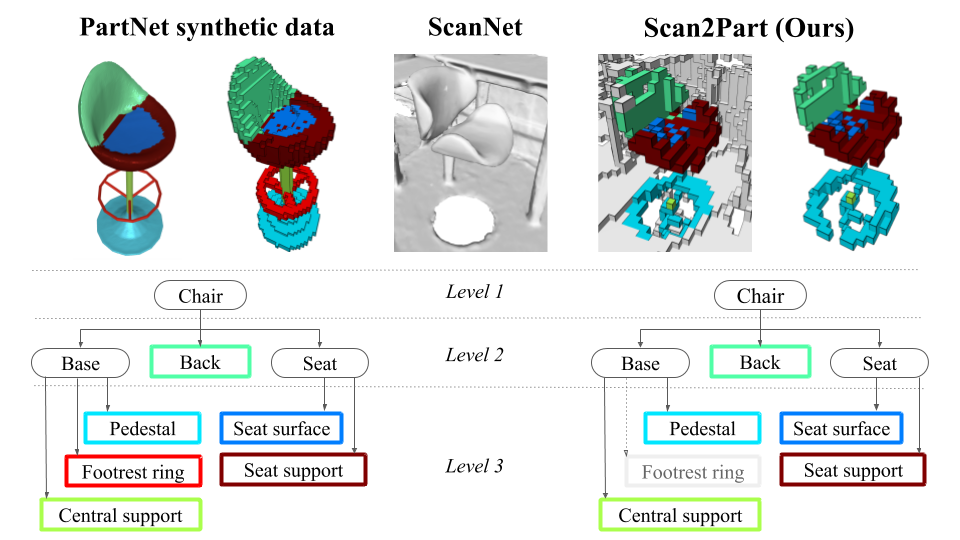}
\caption{Top: our dataset is obtained by combining PartNet synthetic data with ScanNet sensor data. Bottom: the PartNet object hierarchy is compressed to include only parts sufficiently well represented in ScanNet data.}
\end{figure*}

\noindent \textbf{Data preparation.} 
Our first objective is to develop a large-scale 3D scene understanding dataset with part-level annotations. 
We use the 3D geometry in 1,506 3D scenes in Scannet dataset~\cite{dai2017scannet} reconstructed from RGB-D scans in the form of truncated Signed Distance Function (SDF) at voxel resolution = 2\,cm and 5\,cm.
For labeling the scan geometry, we use the parts taxonomy in PartNet dataset~\cite{mo2019partnet} represented as a tree structure where nodes encode parts at various detail levels and edges encode the ``part of'' relationship.
We associate to each 3D scene a per-voxel mask storing leaf part IDs from the taxonomy (i.e., the most fine-grained categories).
To label the 3D scene at a given level~$d$ of semantic detail ($d = 1$ meaning whole objects and $d = 8$ meaning finest parts), we start with the leaf labels in each voxel and traverse the taxonomy tree until hitting depth~$d$.

We further describe the main steps taken to create our Scan2Part benchmark below, leaving the detailed discussion of the technicalities for the supplementary. 
Annotation schema for our dataset is shown in Figure~\ref{fig:dataset_schema}.
The final result of our procedure is a collection of scenes comprised of volumetric instances of separate objects, where background and non-object information is removed, allowing to focus on part based segmentation.

\begin{figure*}[ht!]
\centering
\includegraphics[trim=0 222 0 0,clip,width=0.9\textwidth]{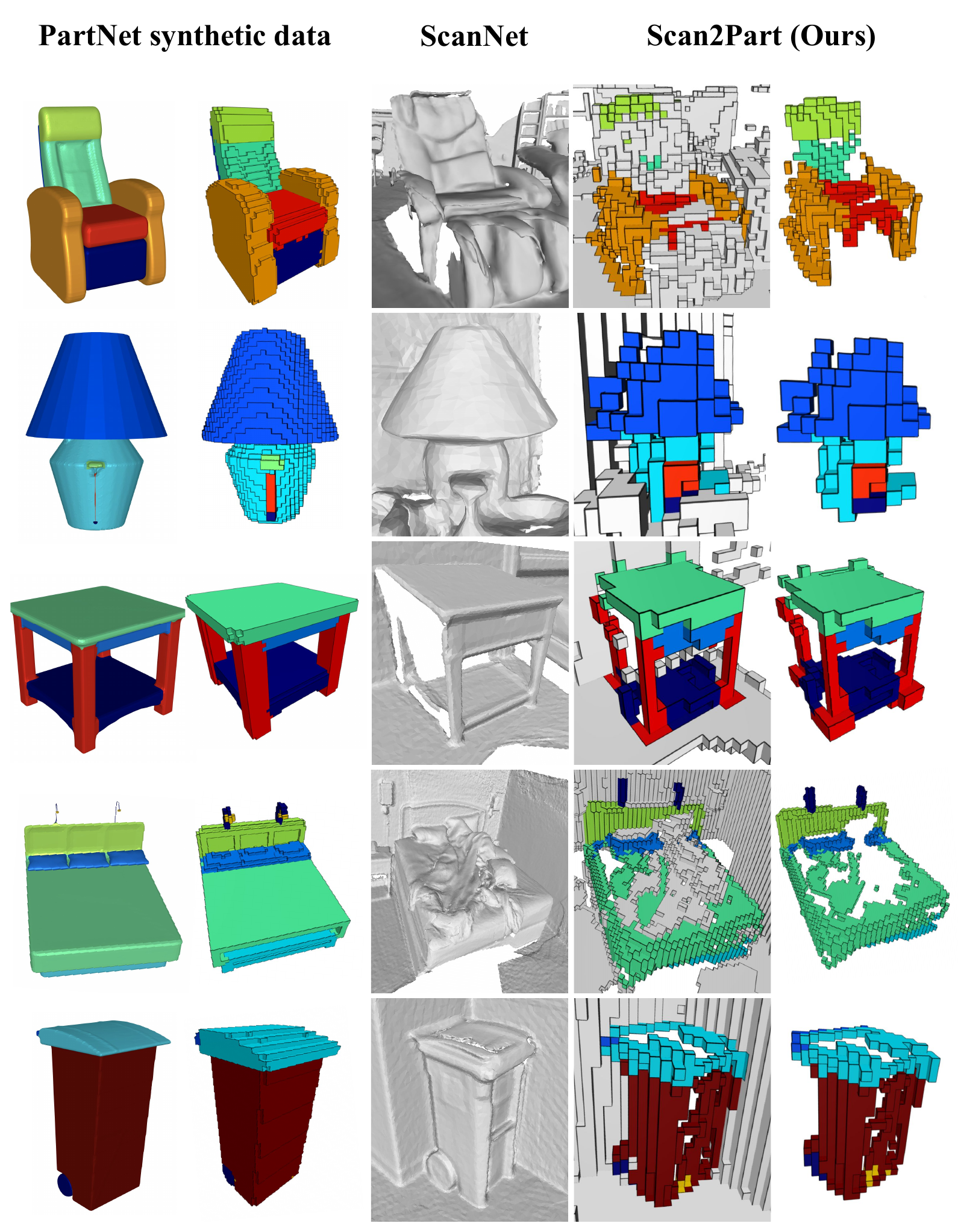}
\caption{Example annotations produced by our automatic procedure at the spatial resolution of 5\,cm$^3$. 
From left to right, part-annotated meshes and voxelized shapes in PartNet; fragments of their respective reconstructions in ScanNet; voxelized part-annotated shapes with and without background (non-object) voxels in Scan2Part.}
\label{fig:dataset_examples}
\end{figure*}

\noindent \textbf{Transferring labels to volumetric 3D grids.} 
To obtain ground-truth semantic parts annotations for real-world 3D scenes, we establish  correspondences between each volume in the 3D scene in Scannet and a set of part-annotated mesh vertices in a registered 3D CAD model from PartNet.
To this end, we first find accurate 9 degrees of freedom (9\,DoF) transformations between PartNet 3D models and their original versions from ShapeNet, and next use the manually annotated 9\,DoF transformations and their respective object categories provided by Scan2CAD to obtain the final scan-to-part alignments.
We further perform a simple majority voting, selecting only the most frequent (among the vertices) part label as the ground truth voxel label.



This procedure results in 242,081 correspondences represented as 9\,DoF transformations between 1,506 reconstructions of real-wold Scannet scenes and 53,618 unique parts of 2,477 ShapeNet objects.
Parts of each object have a tree structure, similar to~\cite{mo2019partnet}.
Note that the majority-based voting implementation of annotation transfer results in some semantic parts labels not being represented in the 3D scan, ultimately affecting parts taxonomy, which we discuss below. 






\noindent \textbf{Parts taxonomy processing. } 
The taxonomy of 3D shape parts is represented as a single tree structure based on~\cite{mo2019partnet}. However, as the voxel resolution of the real-world 3D scan is relatively coarse, particular part categories (e.g., small shape details such as keyboard buttons or door handles) cannot be represented by sufficient number of 3D data points, thus implying a reduction in the original taxonomy. 
We proceed with this reduction by first choosing an appropriate occurrence threshold (we pick 1800 voxels, but demonstrate the effect of different threshold values in the supplementary) and remove classes that have smaller number of representatives in the dataset.
We finalize the taxonomy by pruning trivial paths in the tree (i.e., if a vertex has only one child, then we delete this vertex by connecting the child and the parent of this vertex), but keeping the leaf labels intact. We display the number of vertices at different granularities in the original and resulting part taxonomy levels in Table~\ref{tab:label_presence}.  
Some parts (leafs in the part taxonomy) are not represented in ScanNet data, so we remove these from the tree. 
Figure~\ref{fig:dataset_examples} demonstrates example annotations produced by our automatic procedure.

\begin{table}[ht]
\centering
\resizebox{1.0\columnwidth}{!}{%
    \begin{tabular}{l cccccccc}
    \toprule
    \multirow{2}{*}{\textbf{Level}} & \textbf{1} & \textbf{2} & \textbf{3} & \textbf{4} & \textbf{5} & \textbf{6} & \textbf{7} & \textbf{8} \\
    & \textbf{(object)} & & & & & & & \textbf{(part)} \\
    \midrule
    Full Taxonomy & 18 & 50 & 133 & 223 & 269 & 302 & 306 & 307 \\
    Pruned Taxonomy & 13 & 36 & 79 & --- & --- & --- & --- & --- \\
    \bottomrule
    \end{tabular}
}
\caption{Number of parts on each tree level.}
\label{tab:label_presence}
\end{table}

To aid reproducibility, we will release all the necessary code to combine datasets with minimal efforts while respecting their licence terms.

\subsection{Evaluation Protocol}
\label{dataset:evaluation_protocol}

Based on our dataset, we propose a novel benchmark for part-level scan 3D understanding, offering three core tasks, namely semantic labeling, hierarchical semantic segmentation, and semantic instance segmentation; we overview these tasks in the context of our datasets below.
As inputs in all tasks we supply the voxelized SDF (with RGB information), representing 3D geometry and appearance of individual objects, already separated from the background voxels in the scene.
Note that our tasks are similar to part-level object understanding but operate on real-world 3D shapes.

\noindent \textbf{Evaluation tasks.}
In~\emph{semantic labeling,} the goal is to associate a set of $n$ semantic part labels~$y_j = (y_j^{d_1}, \ldots, y_j^{d_n})$ with each voxel~$v_j$, at detail levels~$d_1, \ldots, d_n$.
Compared to object-level segmentation, the part-level task becomes even more challeging, particularly at deeper levels in the taxonomy that require predicting at increasingly more fine-grained categories.
We provide $1,506$ scenes with $53,618$ unique parts of $2,477$ objects.

For \emph{hierarchical semantic segmentation,} one must perform segmentation at all levels in the hierarchy, inferring labels in coarse- and fine-grained detail levels simultaneously. 

For \emph{instance segmentation,} the goal is to simultaneously perform part-level semantic labeling and assign each voxel~$v_j$ a unique part instance ID (e.g., to differentiate separate legs of a table).

For both semantic and instance segmentations we follow original train/val division of Scannet dataset~\cite{dai2017scannet}.

\begin{figure*}[t]
\label{fig:dataset_stata_1}
\centering
\includegraphics[width=0.95\textwidth]{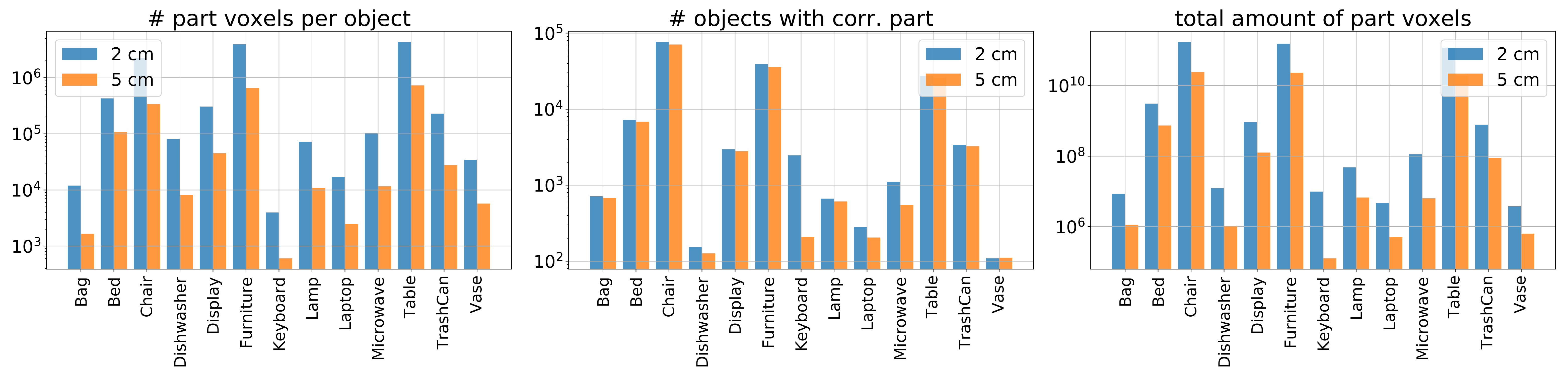}
\caption{Histograms of object statistics for the coarsest ($d = 1$) level of details: LEFT: number of voxels per object, CENTER: the number of corresponding objects in Scan2Part, RIGHT: the total number of object voxels in Scan2Part}
\end{figure*}

\paragraph{The choice of scene understanding levels. }
\label{results:levels}
We evaluate the algorithms at three granularity levels for each object category: coarse-, middle- and fine-grained, roughly corresponding to evaluation in~\cite{mo2019partnet}. 



\paragraph{Quality measures. }
\label{results:metrics}
We evaluate semantic labeling and hierarchical segmentation models by inferring the semantic labels for entire input scenes and computing quality measures at each scene understanding level $d_k$ separately. 
More specifically, for each class $c$ present in the set of classes $\mathbb{C}_k$ at granularity $d_k$, we compute the standard Intersection over Union score $\text{IoU}_c$ and the balanced accuracy score $\text{Acc}_c$.
We report these per-class numbers along with mean IoU and mean balanced accuracy averaged over $\mathbb{C}_k$: $\text{mIoU}_k = \sfrac{1}{n_k}\sum_{c \in \mathbb{C}_k} \text{IoU}_c$, $\text{mAcc}_k = \sfrac{1}{n_k}\sum_{c \in \mathbb{C}_k} \text{Acc}_c$.

We additionally evaluate hierarchical semantic segmentation by averaging mIoU over all hierarchy levels $k \in \{1, \ldots, K\}$.

Instance segmentation is assessed as object detection and thus evaluate this task using average precision (AP) with IoU threshold at 0.5. To generate object hypotheses, each instance is checked against a threshold of confidence equal to 0.25, to filter out noisy voxels.




\subsection{Proposed Approach}
\label{methods:scan2part}

The input to all our models is the voxelized object SDF, representing 3D geometry either with or without associated RGB values.



\noindent \textbf{Part-level 3D understanding.}
To predict parts in our multi-label formulation, we produce a set of softmax scores $p_j = (p_j^{d_1}, \ldots, p_j^{d_n})$.
Our models for semantic and instance segmentation tasks are 3D CNNs identical in architecture up to the last layer. 
We note that training a 3D CNN model for segmentation is a computationally challenging problem; thus, we opted to make it more tractable by using frameworks for sparse differentiable computations. 
Specifically, we use Minkowski Networks~\cite{choy20194d,choy2019fully}, a popular sparse CNN backbone, to implement large sparse fully-convolutional neural networks for geometric feature learning.
We use the Res16UNet34C architecture that showed state-of-the-art performance on multiple scene understanding tasks.
For each input voxel, the network produces a 32-dimensional feature vector that is further appropriately transformed by the last layer, accommodating a required number of predicted classes. All our networks are fully convolutional, enabling inference for scenes with arbitrary spatial extents.




We train the network in multiple setups, differing by the structure of supervision available to the network, each time evaluating labeling performance at each detail level $d_i$. Specifically, we define 
our loss function $L$ to be a weighted sum of cross entropy losses for each level of detail
\begin{equation}
\label{eq:semseg_loss}
L(p, y) = \sum_{k = 1}^K \alpha_k L_{\text{CE}}(p^k, y^k)
\end{equation}
and specify a set of weighting schemes for $\alpha_1, \ldots, \alpha_K$. 
This loss structure allows expressing both ``flat'' segmentation formulations (e.g., choosing $\alpha_{k} = \delta_{ki}$ to segment at level $d_i$ only), that we view as baselines, and multi-task formulations that integrate training signal across multiple levels of semantic detail.

Similarily to~\cite{mo2019partnet}, we approach this task using \emph{bottom-up,} and \emph{top-down} methods. 
Bottom-up method performs segmentation at the most fine-grained level and propagates the labels to object level, leveraging the taxonomy structure.
Conversely, the top-down approach infers labels first at coarse level (starting with objects) and subsequently at finer levels (parts), recursively descending along predicted taxonomy branches.
We note that \emph{multi-task training} using~\eqref{eq:semseg_loss} also results in a hierarchical segmentation method, and include it in the comparison.

\noindent \textbf{Part instance segmentation.} We employ a discriminative loss function which has demonstrated its effectiveness in previous works~\cite{de2017semantic,pham2019jsis3d} and integrates intra-instance clustering and inter-instance separation terms along with a small regularization component.
Compared to architectures that use region proposal modules~\cite{yi2019gspn,pham2019jsis3d,engelmann20203d}, this results in a more computationally lightweight architecture, while reducing instance segmentation problem to a metric learning task.
At inference time, we cluster voxel feature vectors to produce part instances in a scene using mean-shift algorithm~\cite{comaniciu2002mean}.

\section{Experiments}
\label{sec:experiments}

\begin{figure*}[h!]
\centering
\begin{subfigure}{.5\textwidth}
  \centering
  \includegraphics[trim=200 150 200 300,clip,width=.95\linewidth]{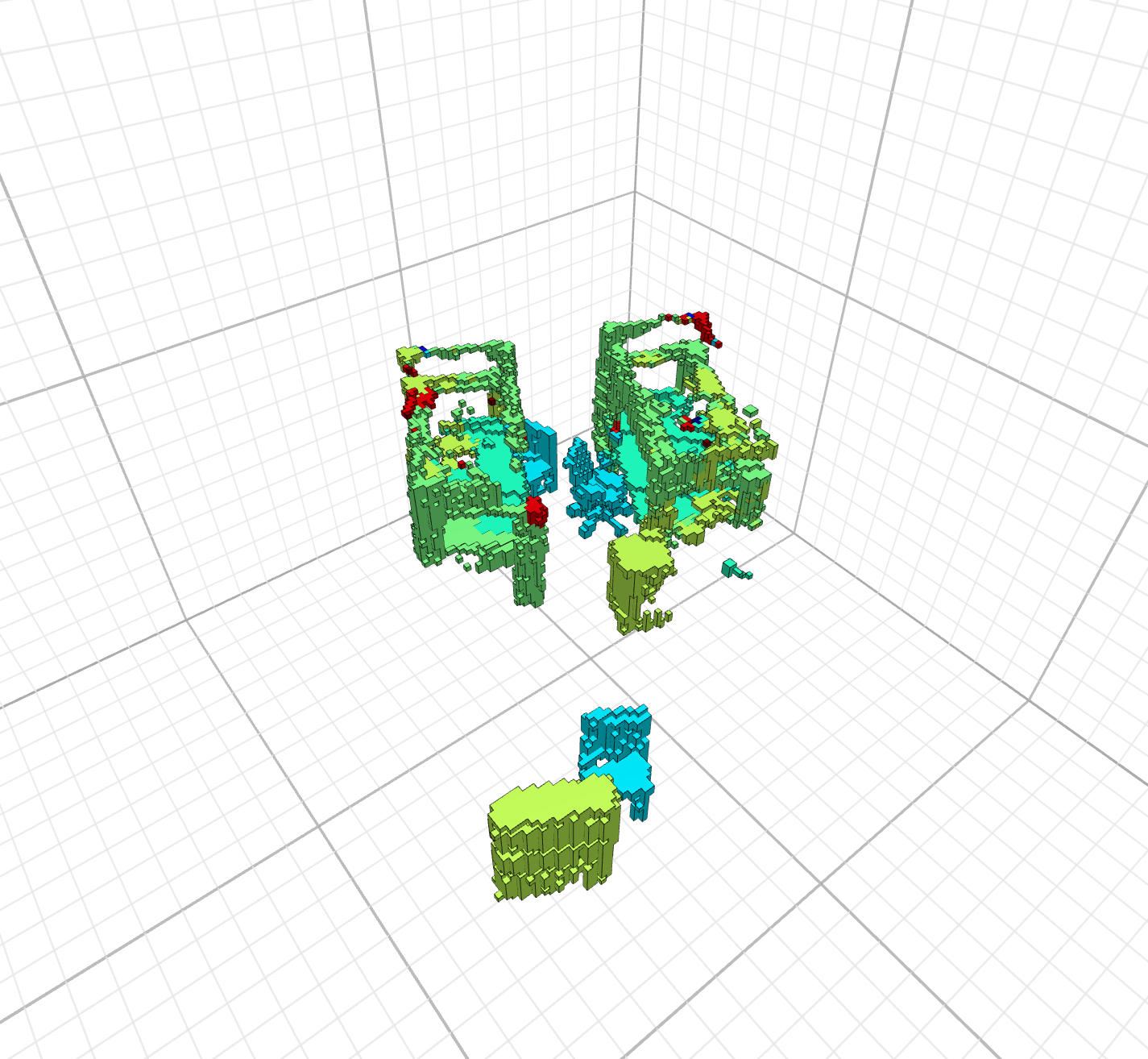}
  \caption{}
  \label{fig:sub11}
\end{subfigure}%
\begin{subfigure}{.5\textwidth}
  \centering
  \includegraphics[trim=200 150 200 300,clip,width=.95\linewidth]{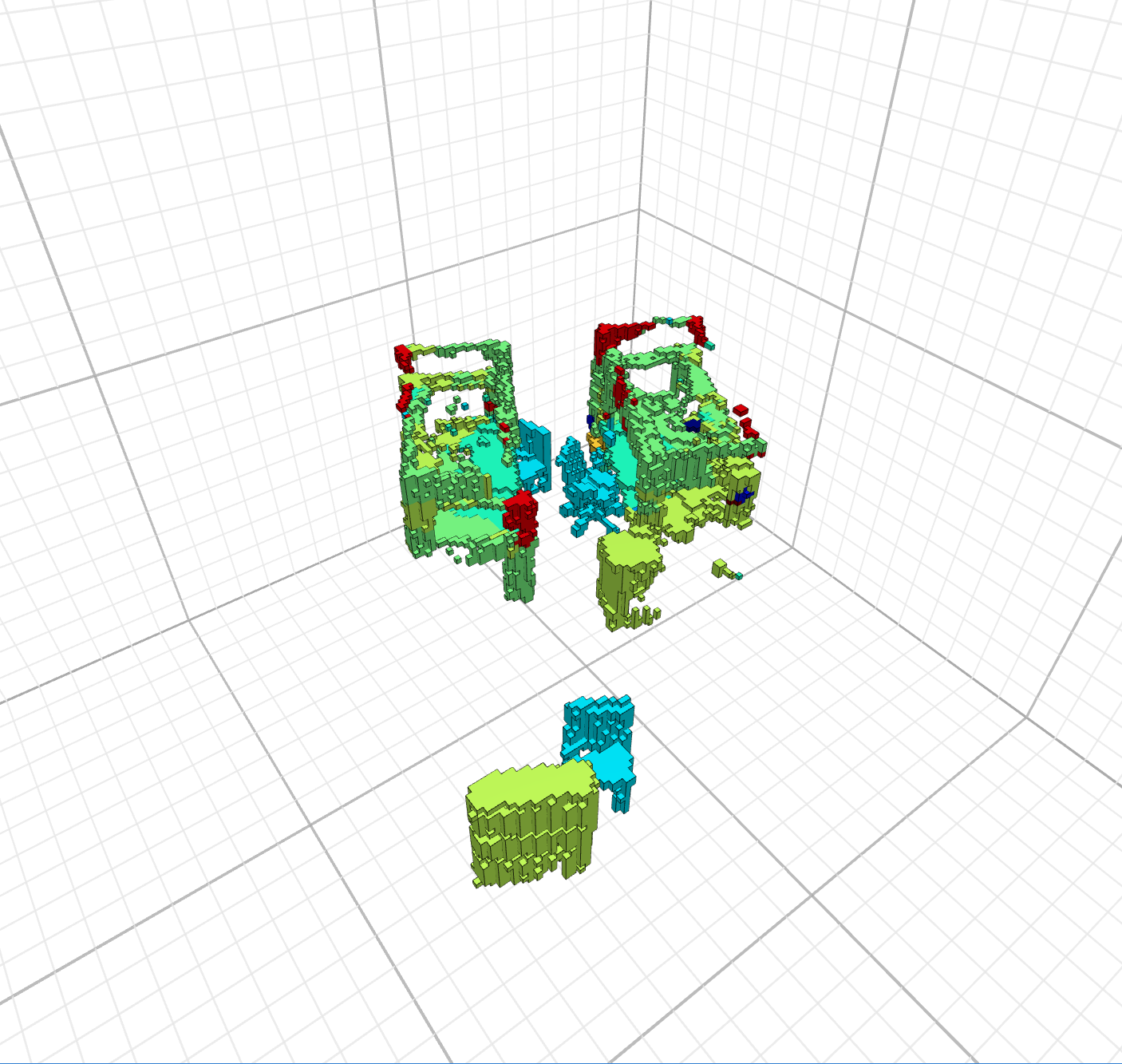}
  \caption{}
  \label{fig:sub22}
\end{subfigure}
\caption{Semantic segmentation prediction from Minkowski Engine (a) and Submanifold (b) models.}
\label{fig:semseg_examples_mink_vs_sub}
\end{figure*}

\begin{table*}[ht]

\centering
\resizebox{.99\textwidth}{!}{
\begin{tabular}{l c ccc ccc ccc ccc c c}  
\toprule
Model &Lvl &Micr. &Disp. &Lamp &Lapt. &Bag &Stor. &Bed &Table &Chair &Dishw. &Trash. &Vase &Keyb. &\textbf{Avg} \\\midrule
Ours (w/ color) & $d_1$ &0.790 &0.676 &0.794 &0.852 &0.808 &0.798 &0.795 &0.852 &0.900 &0.884 &0.840 &1.000 &0.874 &0.839 \\
           & $d_2$ &0.856 &0.356 &0.667 &0.710 &0.275 &0.414 &0.505 &0.467 &0.562 &0.423 &0.640 &0.488 &0.440 &0.757 \\
           & $d_3$ &0.228 &0.146 &0.200 &0.234 &0.484 &0.279 &0.743 &0.667 &0.490 &0.176 &0.333 &0.244 &0.807 &0.711 \\
\midrule
num. instances&       &142 &26 &11 &33 &496 &69 &543 &754 &6 &199 &9 &18 &21 &179 \\
\bottomrule
\end{tabular}
}
\caption{Instance segmentation results on levels $d_1$, $d_2$, and $d_3$}
\label{tab:tbl_instseg}
\end{table*}

\begin{table*}[ht]
\centering

\resizebox{.99\textwidth}{!}{
\begin{tabular}{lrrrrrrrrrrrrrrrr}\toprule
Model &Lvl &Micr. &Disp. &Lamp &Lapt. &Bag &Stor. &Bed &Table &Chair &Dishw. &Trash. &Vase &Keyb. &\textbf{Avg} \\\midrule
Flat &d1 &\textbf{0.408} &0.704 &\textbf{0.558} &0.202 &0.581 &\textbf{0.811} &\textbf{0.777} &\textbf{0.783} &\textbf{0.896} &0.003 &\textbf{0.639} &\textbf{0.403} &0.112 & \textbf{0.529} \\
&d2 &0.199 &0.471 &0.526 &0.298 &0.346 &0.035 &0.423 &0.351 &0.301 &0.183 &0.199 &0.302 &0.033 &0.282 \\
&d3 &0.070 &0.262 &0.380 &0.170 &0.214 &0.213 &0.231 &0.344 &0.193 &0.000 &0.102 &0.208 &0.029 &0.186 \\\midrule
Top-Down &d1 &0.099 &0.266 &0.079 &0.091 &0.338 &0.431 &0.281 &0.527 &0.610 &0.009 &0.276 &0.054 &0.019 &0.237 \\
&d2 &- &\textbf{0.769} &0.442 &\textbf{0.856} &\textbf{0.661} &0.499 &0.339 &0.391 &0.171 &\textbf{0.697} &0.313 &0.199 &\textbf{0.585} &0.494 \\
&d3 &- &0.522 &0.631 &0.439 &0.529 &0.288 &- &0.741 &0.355 &0.682 &0.402 &- &- &0.510 \\\midrule
Bottom-Up &d1 &0.377 &0.626 &0.489 &0.170 &0.515 &0.792 &0.673 &0.748 &0.872 &0.001 &0.582 &0.281 &0.062 &0.476 \\
&d2 &0.188 &0.469 &0.000 &0.041 &0.155 &0.053 &0.006 &0.126 &0.147 &0.000 &0.064 &0.063 &0.031 &0.102 \\
&d3 &0.279 &0.542 &0.198 &0.385 &0.260 &0.232 &0.110 &0.404 &0.335 &0.145 &0.267 &0.128 &0.034 &0.210 \\\midrule
\end{tabular}
}
\caption{Hierarchical semantic segmentation results in terms of mean IoU and mean balanced accuracy for different semantic granularities. We include mIoU averaged over all hierarchy levels as an integral measure.}
\label{tab:tbl_hierarchical_semseg}
\end{table*}

\subsection{Experimental Setup}
\label{exper:setup}

\begin{table*}[ht]

\centering
\resizebox{.80\textwidth}{!}{
\begin{tabular}{l ccc c ccc}
\toprule
 & \multicolumn{3}{c}{Categ. mIoU,\,\% $\uparrow$} && \multicolumn{3}{c}{Inst. mIoU,\,\% $\uparrow$} \\
\cmidrule{2-4}
\cmidrule{6-8}
Method          & $d_1$ & $d_2$ & $d_3$ && $d_1$ & $d_2$ & $d_3$ \\
\midrule
Dense 3D CNN~\cite{lee2017superhuman}  & 0.225 & 0.131 & 0.058 &&  - & - & -  \\
SparseConvNet~\cite{3DSemanticSegmentationWithSubmanifoldSparseConvNet} & 0.4212 & 0.2561 & 0.1789  && 0.7751 & 0.7151 & 0.4923 \\
\midrule
Ours (w/o color)& 0.5179  & 0.2913 & 0.2185  &&  0.8387 & 0.7219 & \textbf{0.5140} \\
Ours (w/ color) & \textbf{0.5290}  & 0.2848 & \textbf{0.2231}  &&  \textbf{0.8422} & 0.7217 & 0.5137 \\
Ours (MTT balanced) & 0.5209 & \textbf{0.3104} & -  &&  0.8364 & \textbf{0.7363} & - \\
Ours (MTT fine-grained) & 0.4953 & 0.2929 & 0.2123 && 0.8065 & 0.7217 & 0.5093 \\
Ours (MTT coarse) & 0.5091 & 0.2963 & 0.1886 && 0.841 & 0.7253 & 0.4414 \\
\midrule
Num. classes       &    13 &    36 &    79 &&    13 &    36 &    79 \\
\bottomrule
\end{tabular}
}
\caption{Fine-grained semantic part segmentation performance in terms of mean IoU and mean per-Instance IoU for different semantic granularities, using our model in various setups and baselines.}
\label{tab:tbl_fine_grained_semseg}
\end{table*}




\noindent \textbf{Data and training.}
We select 80\% of the scenes in ScanNet for training, keeping $5\%$ as a mini-validation to tune hyperparameters, and put aside 20\% scenes for testing. Due to the high imbalance some classes can be under-represented in a testset. Testset was selected through iterative discrete optimization, using desired 80/20 proportion through all classes (especially smaller ones).

We optimize our models using Adam~\cite{kingma2014adam}, initializing learning rate proportionally to the batch size with base learning rate = 0.003 and $\beta_1=0.9$, $\beta_2=0.999$. Learning rate during training follows Multi-step exponential decay schedule with $\gamma=0.2$, and a weight decay of $10^{-4}$. We train until early stopping with patience parameter of 10, or until we reach a maximum epoch limit of 200. Depending on the specific model training can take several hours on our dedicated server equiped with 4 Nvidia GTX 1080.

\noindent \textbf{Baselines.}
SparseConvNet~\cite{3DSemanticSegmentationWithSubmanifoldSparseConvNet} is an implementation of sparse 3D CNNs; though it has a smaller number of sparse layer types, it has enough features to implement a U-Net style fully-convolutional architecture we need to compute geometric features of voxels.
As a baseline, we also train a Dense 3D CNN~\cite{lee2017superhuman} model based on U-Net architecture with ResNet blocks.

To train the baselines, we extract 16~volumetric crops of size~$64^3$ from each voxelized scene and randomly shuffle them across all scenes in the training split.
The batch size was selected from 4 to 48 in different experiments, subject to fitting the model into the available GPU memory.
Because this model's implementation requires slicing of the scene, it cannot be evaluated for per-instance metrics.


\noindent \textbf{Semantic Part Segmentation.}
\label{results:model_configs}
Table~\ref{table:configurations} summarizes our model specifications, differing by the choice of weights in~\eqref{eq:semseg_loss}. The first three rows correspond to training a single level-of-detail semantic segmentation model. The last three rows define a Multi-Task-Training (MTT) objective where learned features of the occupied voxels are projected using linear layers to different level labels, and their loss functions are combined for training. 

\begin{table}[h]
\centering
\begin{tabular}{lrrr}\\\toprule  
Configuration   & $\alpha_1$ & $\alpha_2$ & $\alpha_3$ \\
\midrule
Base coarse     & 1 & 0 & 0 \\
Base middle     & 0 & 1 & 0 \\
Base fine       & 0 & 0 & 1 \\
MTT-12          & .5 & .5 &  0 \\
MTT-123-coarse  & .7 & .2 & .1 \\
MTT-123-fine    & .1 & .2 & .7 \\
\bottomrule
\end{tabular}
\caption{Our objective configurations in~\eqref{eq:semseg_loss}.}
\label{table:configurations}
\end{table}

\noindent \textbf{Part Instance Segmentation.}
This task test the ability of our models to separate parts solely based on their shape and mutual position within object, disregarding semantic class information.

\noindent \textbf{Hierarchical Segmentation.}
Performing hierarchical segmentation of voxels in a scene according to a taxonomy of objects and their parts described in~\ref{dataset:construction} is a challenging task that can be approached in different ways:

\begin{itemize}
\item \textbf{Top-Down} approach assigns labels to a voxel by solving a sequence of smaller classification tasks. This approach predicts the distribution of semantic object labels first, followed by prediction of first-level parts, and continuing until a voxel can be assigned a leaf label from the taxonomy. Every prediction produces SoftMax distribution over possible sub-parts; voxels are masked based on the ground truth of the "parent" part during evaluation.

\item \textbf{Bottom-Up} A model predicts a leaf label to a voxel on the finest level-of-detail, and semantic part segmentation on any level-of-detail can be computed by "projecting" up. Probabilities of part labels having the same "parent" part are added together.

\end{itemize}

\subsection{Comparative Studies}
\label{exper:comparative}

\noindent \textbf{Can we recognize parts in real-world scans?}
Table~\ref{tab:tbl_fine_grained_semseg} demonstrates part-level semantic 3D understanding performance of our approach vs. the baseline methods. We show that adding semantic information from multiple levels of detail via part objectives in in~\eqref{eq:semseg_loss} significantly improves models segmentation performance across hierarchy levels.

\begin{figure}[t]

\centering
\includegraphics[trim=40 20 40 150,clip,width=0.48\textwidth]{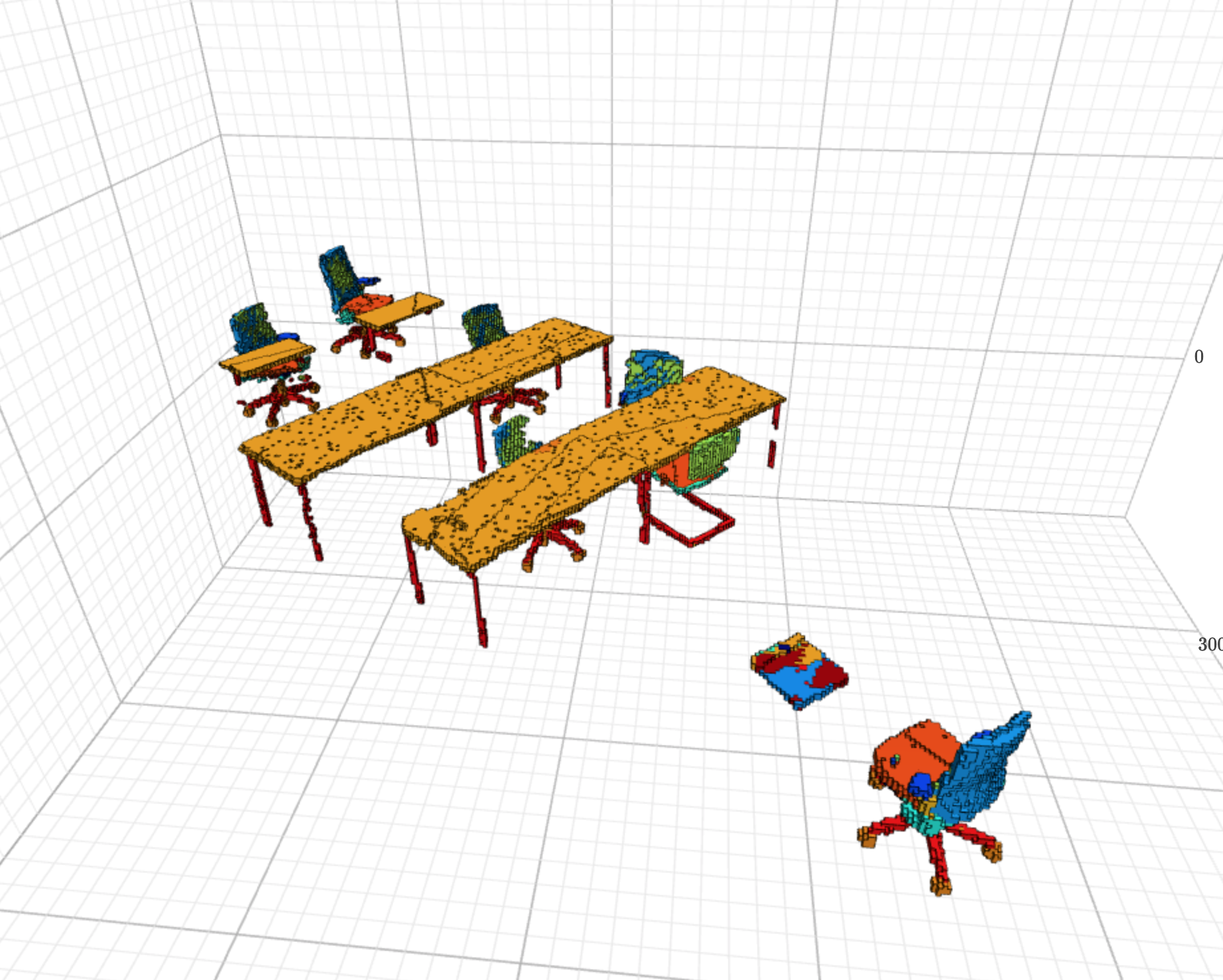}
\caption{Qualitative hierarchical semantic segmentation results using our models, trained in each respective part-category level and our proposed method. Note the segmentation performance, particularly at finer levels in the parts taxonomy}

\label{fig:fig_fine_grained_semseg}

\end{figure}

\noindent \textbf{Does adding part-level annotation improve object-level 3D understanding?}
Table~\ref{tab:tbl_hierarchical_semseg} demonstrates results across different object classes and in different inference setups obtained at the object-level and lower level-of-detail.

\noindent \textbf{Does pre-training for part segmentation help achieve hierarchical / instance tasks?}
We present performance of hierarchical semantic segmentation in Table~\ref{tab:tbl_hierarchical_semseg}.
Note that despite the baselines are focusing solely on a single level of semantic detail, our method is able to leverage a multi-task objective in~\eqref{eq:semseg_loss} to perform more efficient segmentation.

\noindent \textbf{Part instance segmentation results. }
\label{results:instance}
We present instance segmentation results in Table~\ref{tab:tbl_instseg}, results indicate that segmentation accuracy decreases when the level of detail is reduced, presumably because object-part-masks become more generic, closet to shape primitives. The metric performs on par to the object level on particular objects with less shape variability (e.g., Bag, Bed, Table, Keyboard).

\subsection{Ablative Studies}
\label{exper:ablative}

\begin{table}[ht]

\centering
\resizebox{.99\columnwidth}{!}{
\begin{tabular}{l ccc c ccc}
\toprule
 & \multicolumn{3}{c}{Categ. mIoU,\,\% $\uparrow$} && \multicolumn{3}{c}{Inst. mIoU,\,\% $\uparrow$} \\
\cmidrule{2-4}
\cmidrule{6-8}
Method               & $d_1$ & $d_2$ & $d_3$ && $d_1$ & $d_2$ & $d_3$ \\
\midrule
Voxel size 5\,cm$^3$ & 0.5179 & 0.2913 & 0.2185  && 0.8387 & 0.7219 & 0.5140 \\
Voxel size 2\,cm$^3$ & 0.6465 & 0.4138 & 0.3415 && 0.8958 &  0.8263 & 0.6295 \\
\midrule
Num. classes       &    13 &    36 &    79 &&    13 &    36 &    79 \\
\bottomrule
\end{tabular}
}
\caption{Our method is able to effectively work on various voxel resolution leves. On 5cm resolution tipical training time for equal number of epochs increased from several hours to a day.}
\label{tab:tbl_voxel_resolution}
\end{table}

\begin{figure*}[h!]
\centering
\begin{subfigure}{.5\textwidth}
  \centering
  \includegraphics[trim=600 200 600 300,clip,width=.99\linewidth]{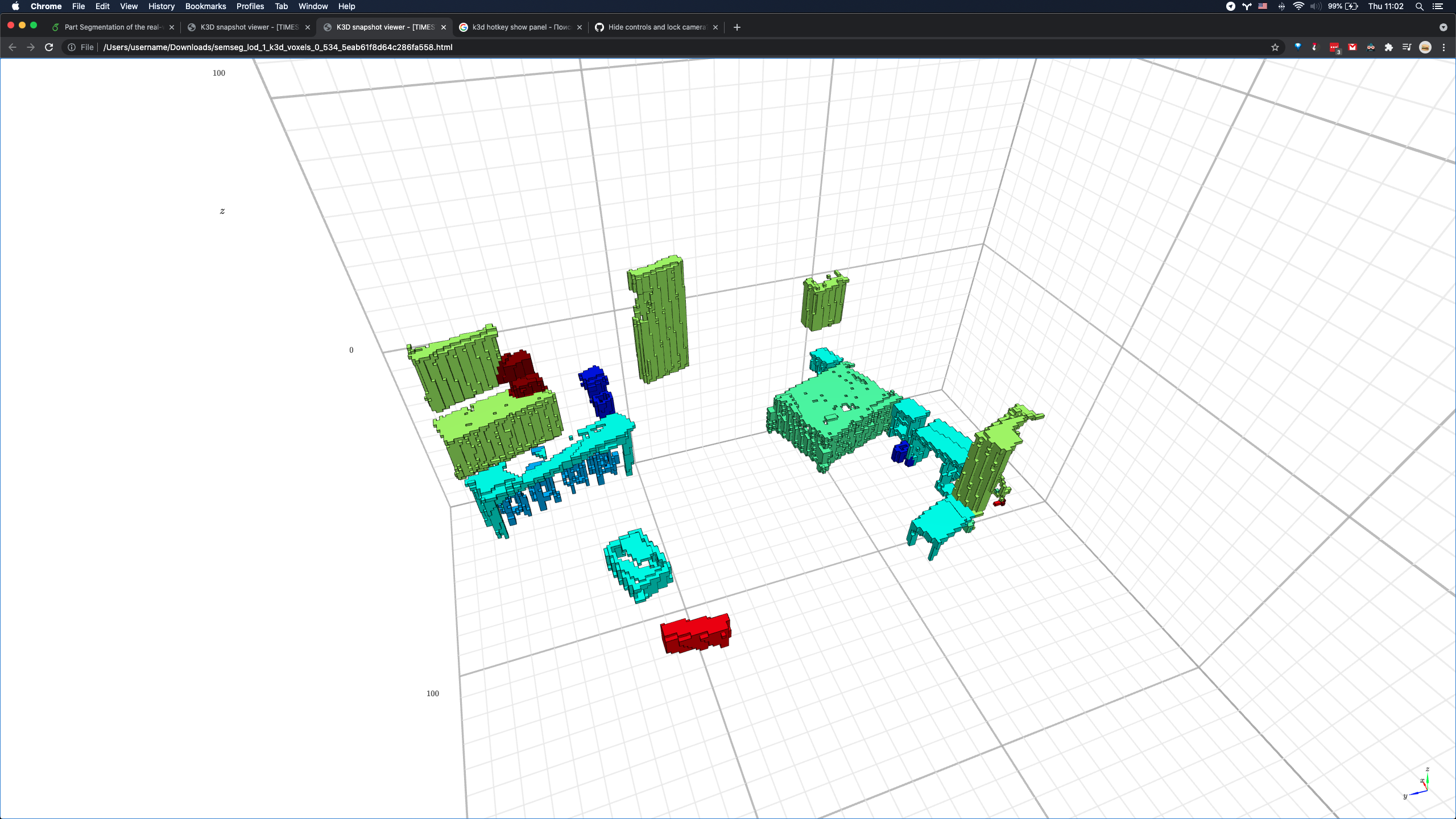}
  \caption{}
  \label{fig:sub1}
\end{subfigure}%
\begin{subfigure}{.5\textwidth}
  \centering
  \includegraphics[trim=600 200 600 300,clip,width=.99\linewidth]{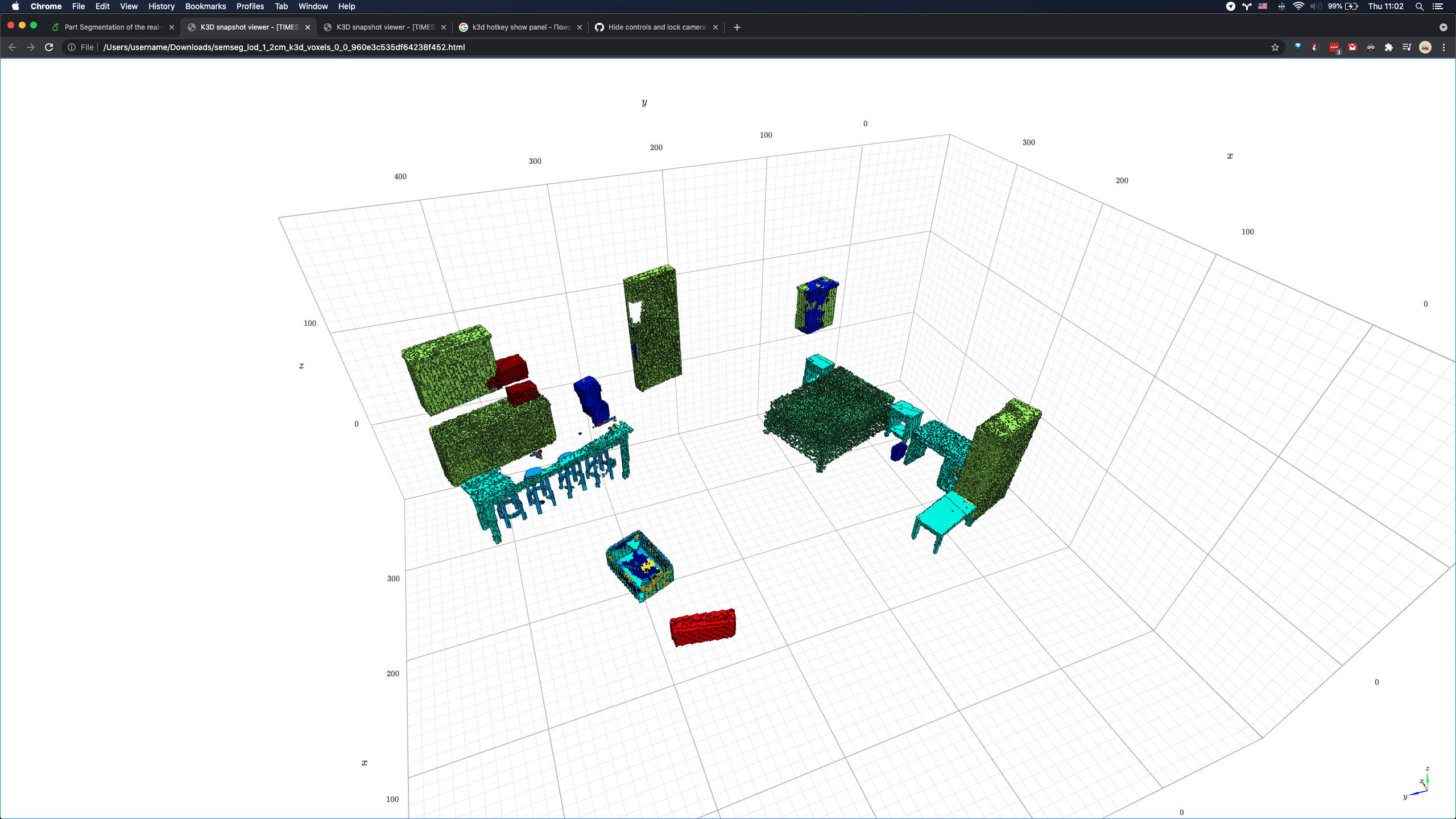}
  \caption{}
  \label{fig:sub2}
\end{subfigure}
\caption{Semantic segmentation results for the same scene voxelized at 5\,cm$^3$ (a) and 2\,cm$^3$ (b) voxel resolution. Despite coarser geometry yields somewhat more robust segmentation, it does not allow representing finer parts.}
\label{fig:semseg_examples_2cm_5cm}
\end{figure*}

What is required for efficient part-level understanding? The following ablations can inform to what is more important in our problem domain.

\noindent \textbf{Effect of backbone network.}
In Table~\ref{tab:tbl_fine_grained_semseg} one can see that having a sparse backbone is crucial to the semantic segmentation ability of the model. There exists a moderate effect between model size/complexity and performance. We are confident that we got close to the limits of segmentation performance given the quality of data in the benchmark.

\noindent \textbf{Effect of loss.}
Most of our models in Table~\ref{tab:tbl_fine_grained_semseg} are trained with weighted cross-entropy Loss to combat high unbalance for part labels in scenes. Training on part segmentation in a multi-task setup with different choices of objective weights has demonstrated a trade-off in the effectiveness of the geometric features and their ability to predict labels on different level-of-details. 

\noindent \textbf{Effect of color information.}
Surprisingly, introducing color information only contributes slightly to the performance of semantic part segmentation (see Table~\ref{tab:tbl_fine_grained_semseg}). This result disagrees with the expectations that voxel color would be a highly predictive factor for part correspondence. We hypothesize that this could be due to the high variability in lightning and acquisition conditions in the original ScanNet dataset.

\noindent \textbf{Effect of voxel resolution.}
We conducted experiments at a $2.5\times$ finer voxel resolution of 2\,cm$^3$ and present their results in Table~\ref{tab:tbl_voxel_resolution}. 
Despite higher complexity geometry, we have observed an improvement in performance metrics across levels of detail.
However, due to a significant increase in computational requirements, we performed the majority of experiments on the 5\,cm$^3$ version of our dataset.

\section{Conclusion}
We introduced Scan2Part, a novel method and a challenging benchmark for part-level understanding of real-world 3D objects.
The core of our method is to leverage structural knowledge of objects composition to perform a variety of segmentation tasks in setting with complex geometry, high levels of uncertainty due to noise. To achieve that, we explore the part taxonomies of common objects in indoor scenes, on multiple scales and methods of compressing them for more effective use in machine learning applications. We demonstrated that specific ways of training deep segmentation models like ours sparse Residual-U-Net architecture on these novel tasks we introduced are better at capturing inductive biases in structured labels on some parts of a taxonomy but not the others. Further research on relationships between structure of real-world scenes and perception models is required and we hope our benchmark and dataset will accelerate it.  
In the near future we plan on releasing a second version of the dataset with background of scenes is not removed, so that and we can study performance of our models in scenarios closer to raw sensor data.


\section{Acknowledgements}

This work has been supported by the Russian Science Foundation under Grant 19-41-04109. 
The authors acknowledge the use of Zhores~\cite{zacharov2019zhores} for obtaining the results presented in this paper.


{\small
\bibliographystyle{apalike}
\bibliography{references}

\begin{thebibliography}{}

\bibitem[{Armeni} et~al., 2017]{2017arXiv170201105A}
{Armeni}, I., {Sax}, A., {Zamir}, A.~R., and {Savarese}, S. (2017).
\newblock {Joint 2D-3D-Semantic Data for Indoor Scene Understanding}.
\newblock {\em ArXiv e-prints}.

\bibitem[Armeni et~al., 2016]{armeni20163d}
Armeni, I., Sener, O., Zamir, A.~R., Jiang, H., Brilakis, I., Fischer, M., and
  Savarese, S. (2016).
\newblock 3d semantic parsing of large-scale indoor spaces.
\newblock In {\em Proceedings of the IEEE Conference on Computer Vision and
  Pattern Recognition}, pages 1534--1543.

\bibitem[Avetisyan et~al., 2019]{avetisyan2019scan2cad}
Avetisyan, A., Dahnert, M., Dai, A., Savva, M., Chang, A.~X., and Nie{\ss}ner,
  M. (2019).
\newblock Scan2cad: Learning cad model alignment in rgb-d scans.
\newblock In {\em Proceedings of the IEEE Conference on Computer Vision and
  Pattern Recognition}, pages 2614--2623.

\bibitem[Biederman, 1987]{biederman1987recognition}
Biederman, I. (1987).
\newblock Recognition-by-components: a theory of human image understanding.
\newblock {\em Psychological review}, 94(2):115.

\bibitem[Bokhovkin et~al., 2021]{bokhovkin2021towards}
Bokhovkin, A., Ishimtsev, V., Bogomolov, E., Zorin, D., Artemov, A., Burnaev,
  E., and Dai, A. (2021).
\newblock Towards part-based understanding of rgb-d scans.
\newblock In {\em Proceedings of the IEEE/CVF Conference on Computer Vision and
  Pattern Recognition}, pages 7484--7494.

\bibitem[Chang et~al., 2017]{chang2017matterport3d}
Chang, A., Dai, A., Funkhouser, T., Halber, M., Niebner, M., Savva, M., Song,
  S., Zeng, A., and Zhang, Y. (2017).
\newblock Matterport3d: Learning from rgb-d data in indoor environments.
\newblock In {\em 2017 International Conference on 3D Vision (3DV)}, pages
  667--676. IEEE.

\bibitem[Chang et~al., 2015]{chang2015shapenet}
Chang, A.~X., Funkhouser, T., Guibas, L., Hanrahan, P., Huang, Q., Li, Z.,
  Savarese, S., Savva, M., Song, S., Su, H., et~al. (2015).
\newblock Shapenet: An information-rich 3d model repository.
\newblock {\em arXiv preprint arXiv:1512.03012}.

\bibitem[Chen et~al., 2009]{Chen:2009:ABF}
Chen, X., Golovinskiy, A., and Funkhouser, T. (2009).
\newblock A benchmark for {3D} mesh segmentation.
\newblock {\em ACM Transactions on Graphics (Proc. SIGGRAPH)}, 28(3).

\bibitem[Chen et~al., 2019]{chen2019bae}
Chen, Z., Yin, K., Fisher, M., Chaudhuri, S., and Zhang, H. (2019).
\newblock Bae-net: branched autoencoder for shape co-segmentation.
\newblock In {\em Proceedings of the IEEE International Conference on Computer
  Vision}, pages 8490--8499.

\bibitem[Choy et~al., 2019a]{choy20194d}
Choy, C., Gwak, J., and Savarese, S. (2019a).
\newblock 4d spatio-temporal convnets: Minkowski convolutional neural networks.
\newblock In {\em Proceedings of the IEEE Conference on Computer Vision and
  Pattern Recognition}, pages 3075--3084.

\bibitem[Choy et~al., 2019b]{choy2019fully}
Choy, C., Park, J., and Koltun, V. (2019b).
\newblock Fully convolutional geometric features.
\newblock In {\em Proceedings of the IEEE International Conference on Computer
  Vision}, pages 8958--8966.

\bibitem[Comaniciu and Meer, 2002]{comaniciu2002mean}
Comaniciu, D. and Meer, P. (2002).
\newblock Mean shift: A robust approach toward feature space analysis.
\newblock {\em IEEE Transactions on pattern analysis and machine intelligence},
  24(5):603--619.

\bibitem[Dai et~al., 2017]{dai2017scannet}
Dai, A., Chang, A.~X., Savva, M., Halber, M., Funkhouser, T., and Nie{\ss}ner,
  M. (2017).
\newblock Scannet: Richly-annotated 3d reconstructions of indoor scenes.
\newblock {\em arXiv preprint arXiv:1702.04405}.

\bibitem[Dai and Nie{\ss}ner, 2018]{dai20183dmv}
Dai, A. and Nie{\ss}ner, M. (2018).
\newblock 3dmv: Joint 3d-multi-view prediction for 3d semantic scene
  segmentation.
\newblock In {\em Proceedings of the European Conference on Computer Vision
  (ECCV)}, pages 452--468.

\bibitem[De~Brabandere et~al., 2017]{de2017semantic}
De~Brabandere, B., Neven, D., and Van~Gool, L. (2017).
\newblock Semantic instance segmentation with a discriminative loss function.
\newblock {\em arXiv preprint arXiv:1708.02551}.

\bibitem[Elich et~al., 2019]{elich20193d}
Elich, C., Engelmann, F., Schult, J., Kontogianni, T., and Leibe, B. (2019).
\newblock 3d-bevis: birds-eye-view instance segmentation.
\newblock {\em arXiv preprint arXiv:1904.02199}.

\bibitem[Engelmann et~al., 2020]{engelmann20203d}
Engelmann, F., Bokeloh, M., Fathi, A., Leibe, B., and Nie{\ss}ner, M. (2020).
\newblock 3d-mpa: Multi proposal aggregation for 3d semantic instance
  segmentation.
\newblock {\em arXiv preprint arXiv:2003.13867}.

\bibitem[Fisher et~al., 2012]{2012-scenesynth}
Fisher, M., Ritchie, D., Savva, M., Funkhouser, T., and Hanrahan, P. (2012).
\newblock Example-based synthesis of 3d object arrangements.
\newblock In {\em ACM SIGGRAPH Asia 2012 papers}, SIGGRAPH Asia '12.

\bibitem[Garcia-Garcia et~al., 2018]{garcia2018robotrix}
Garcia-Garcia, A., Martinez-Gonzalez, P., Oprea, S., Castro-Vargas, J.~A.,
  Orts-Escolano, S., Garcia-Rodriguez, J., and Jover-Alvarez, A. (2018).
\newblock The robotrix: An extremely photorealistic and very-large-scale indoor
  dataset of sequences with robot trajectories and interactions.
\newblock In {\em 2018 IEEE/RSJ International Conference on Intelligent Robots
  and Systems (IROS)}, pages 6790--6797. IEEE.

\bibitem[Graham et~al.,
  2018]{3DSemanticSegmentationWithSubmanifoldSparseConvNet}
Graham, B., Engelcke, M., and van~der Maaten, L. (2018).
\newblock 3d semantic segmentation with submanifold sparse convolutional
  networks.
\newblock {\em CVPR}.

\bibitem[Handa et~al., 2016]{handa2016understanding}
Handa, A., Patraucean, V., Badrinarayanan, V., Stent, S., and Cipolla, R.
  (2016).
\newblock Understanding real world indoor scenes with synthetic data.
\newblock In {\em Proceedings of the IEEE Conference on Computer Vision and
  Pattern Recognition}, pages 4077--4085.

\bibitem[Hoffman and Richards, 1984]{hoffman1984parts}
Hoffman, D.~D. and Richards, W.~A. (1984).
\newblock Parts of recognition.
\newblock {\em Cognition}, 18(1-3):65--96.

\bibitem[Hou et~al., 2019]{hou20193d}
Hou, J., Dai, A., and Nie{\ss}ner, M. (2019).
\newblock 3d-sis: 3d semantic instance segmentation of rgb-d scans.
\newblock In {\em Proceedings of the IEEE Conference on Computer Vision and
  Pattern Recognition}, pages 4421--4430.

\bibitem[Hua et~al., 2016]{hua2016scenenn}
Hua, B.-S., Pham, Q.-H., Nguyen, D.~T., Tran, M.-K., Yu, L.-F., and Yeung,
  S.-K. (2016).
\newblock Scenenn: A scene meshes dataset with annotations.
\newblock In {\em 2016 Fourth International Conference on 3D Vision (3DV)},
  pages 92--101. IEEE.

\bibitem[Johnson et~al., 2018]{DBLP:journals/corr/abs-1804-01622}
Johnson, J., Gupta, A., and Fei{-}Fei, L. (2018).
\newblock Image generation from scene graphs.
\newblock {\em CoRR}, abs/1804.01622.

\bibitem[Keselman et~al., 2017]{keselman2017intel}
Keselman, L., Iselin~Woodfill, J., Grunnet-Jepsen, A., and Bhowmik, A. (2017).
\newblock Intel realsense stereoscopic depth cameras.
\newblock In {\em Proceedings of the IEEE Conference on Computer Vision and
  Pattern Recognition Workshops}, pages 1--10.

\bibitem[Kingma and Ba, 2014]{kingma2014adam}
Kingma, D.~P. and Ba, J. (2014).
\newblock Adam: A method for stochastic optimization.
\newblock {\em arXiv preprint arXiv:1412.6980}.

\bibitem[Klokov and Lempitsky, 2017]{klokov2017escape}
Klokov, R. and Lempitsky, V. (2017).
\newblock Escape from cells: Deep kd-networks for the recognition of 3d point
  cloud models.
\newblock In {\em Proceedings of the IEEE International Conference on Computer
  Vision}, pages 863--872.

\bibitem[Krishna et~al., 2017]{krishna2017visual}
Krishna, R., Zhu, Y., Groth, O., Johnson, J., Hata, K., Kravitz, J., Chen, S.,
  Kalantidis, Y., Li, L.-J., Shamma, D.~A., et~al. (2017).
\newblock Visual genome: Connecting language and vision using crowdsourced
  dense image annotations.
\newblock {\em International Journal of Computer Vision}, 123(1):32--73.

\bibitem[Lahoud et~al., 2019]{lahoud20193d}
Lahoud, J., Ghanem, B., Pollefeys, M., and Oswald, M.~R. (2019).
\newblock 3d instance segmentation via multi-task metric learning.
\newblock In {\em Proceedings of the IEEE International Conference on Computer
  Vision}, pages 9256--9266.

\bibitem[Lee et~al., 2017]{lee2017superhuman}
Lee, K., Zung, J., Li, P., Jain, V., and Seung, H.~S. (2017).
\newblock Superhuman accuracy on the snemi3d connectomics challenge.
\newblock {\em arXiv preprint arXiv:1706.00120}.

\bibitem[Li et~al., 2017]{li2017grass}
Li, J., Xu, K., Chaudhuri, S., Yumer, E., Zhang, H., and Guibas, L. (2017).
\newblock Grass: Generative recursive autoencoders for shape structures.
\newblock {\em ACM Transactions on Graphics (TOG)}, 36(4):52.

\bibitem[Li et~al., 2018]{InteriorNet18}
Li, W., Saeedi, S., McCormac, J., Clark, R., Tzoumanikas, D., Ye, Q., Huang,
  Y., Tang, R., and Leutenegger, S. (2018).
\newblock Interiornet: Mega-scale multi-sensor photo-realistic indoor scenes
  dataset.
\newblock In {\em British Machine Vision Conference (BMVC)}.

\bibitem[Liang et~al., 2019]{liang20193d}
Liang, Z., Yang, M., and Wang, C. (2019).
\newblock 3d graph embedding learning with a structure-aware loss function for
  point cloud semantic instance segmentation.
\newblock {\em arXiv preprint arXiv:1902.05247}.

\bibitem[Liu and Furukawa, 2019]{liu2019masc}
Liu, C. and Furukawa, Y. (2019).
\newblock Masc: multi-scale affinity with sparse convolution for 3d instance
  segmentation.
\newblock {\em arXiv preprint arXiv:1902.04478}.

\bibitem[McCormac et~al., 2017]{McCormac:etal:ICCV2017}
McCormac, J., Handa, A., Leutenegger, S., and J.Davison, A. (2017).
\newblock Scenenet rgb-d: Can 5m synthetic images beat generic imagenet
  pre-training on indoor segmentation?

\bibitem[Mo et~al., 2019a]{mo2019structurenet}
Mo, K., Guerrero, P., Yi, L., Su, H., Wonka, P., Mitra, N., and Guibas, L.~J.
  (2019a).
\newblock Structurenet: Hierarchical graph networks for 3d shape generation.
\newblock {\em arXiv preprint arXiv:1908.00575}.

\bibitem[Mo et~al., 2020]{mo2020pt2pc}
Mo, K., Wang, H., Yan, X., and Guibas, L.~J. (2020).
\newblock Pt2pc: Learning to generate 3d point cloud shapes from part tree
  conditions.
\newblock {\em arXiv preprint arXiv:2003.08624}.

\bibitem[Mo et~al., 2019b]{mo2019partnet}
Mo, K., Zhu, S., Chang, A.~X., Yi, L., Tripathi, S., Guibas, L.~J., and Su, H.
  (2019b).
\newblock Partnet: A large-scale benchmark for fine-grained and hierarchical
  part-level 3d object understanding.
\newblock In {\em Proceedings of the IEEE Conference on Computer Vision and
  Pattern Recognition}, pages 909--918.

\bibitem[Pham et~al., 2019a]{pham2019real}
Pham, Q.-H., Hua, B.-S., Nguyen, T., and Yeung, S.-K. (2019a).
\newblock Real-time progressive 3d semantic segmentation for indoor scenes.
\newblock In {\em 2019 IEEE Winter Conference on Applications of Computer
  Vision (WACV)}, pages 1089--1098. IEEE.

\bibitem[Pham et~al., 2019b]{pham2019jsis3d}
Pham, Q.-H., Nguyen, T., Hua, B.-S., Roig, G., and Yeung, S.-K. (2019b).
\newblock Jsis3d: joint semantic-instance segmentation of 3d point clouds with
  multi-task pointwise networks and multi-value conditional random fields.
\newblock In {\em Proceedings of the IEEE Conference on Computer Vision and
  Pattern Recognition}, pages 8827--8836.

\bibitem[Qi et~al., 2017a]{qi2017pointnet}
Qi, C.~R., Su, H., Mo, K., and Guibas, L.~J. (2017a).
\newblock Pointnet: Deep learning on point sets for 3d classification and
  segmentation.
\newblock In {\em Proceedings of the IEEE conference on computer vision and
  pattern recognition}, pages 652--660.

\bibitem[Qi et~al., 2017b]{qi2017pointnet++}
Qi, C.~R., Yi, L., Su, H., and Guibas, L.~J. (2017b).
\newblock Pointnet++: Deep hierarchical feature learning on point sets in a
  metric space.
\newblock {\em arXiv preprint arXiv:1706.02413}.

\bibitem[Russell et~al., 2009]{russell2009associative}
Russell, C., Kohli, P., Torr, P.~H., et~al. (2009).
\newblock Associative hierarchical crfs for object class image segmentation.
\newblock In {\em Computer Vision, 2009 IEEE 12th International Conference on},
  pages 739--746. IEEE.

\bibitem[Song et~al., 2017]{song2016ssc}
Song, S., Yu, F., Zeng, A., Chang, A.~X., Savva, M., and Funkhouser, T. (2017).
\newblock Semantic scene completion from a single depth image.
\newblock {\em Proceedings of 30th IEEE Conference on Computer Vision and
  Pattern Recognition}.

\bibitem[Straub et~al., 2019]{replica19arxiv}
Straub, J., Whelan, T., Ma, L., Chen, Y., Wijmans, E., Green, S., Engel, J.~J.,
  Mur-Artal, R., Ren, C., Verma, S., Clarkson, A., Yan, M., Budge, B., Yan, Y.,
  Pan, X., Yon, J., Zou, Y., Leon, K., Carter, N., Briales, J., Gillingham, T.,
  Mueggler, E., Pesqueira, L., Savva, M., Batra, D., Strasdat, H.~M., Nardi,
  R.~D., Goesele, M., Lovegrove, S., and Newcombe, R. (2019).
\newblock The {R}eplica dataset: A digital replica of indoor spaces.
\newblock {\em arXiv preprint arXiv:1906.05797}.

\bibitem[Sun et~al., 2019]{sun2019learning}
Sun, C.-Y., Zou, Q.-F., Tong, X., and Liu, Y. (2019).
\newblock Learning adaptive hierarchical cuboid abstractions of 3d shape
  collections.
\newblock {\em ACM Transactions on Graphics (TOG)}, 38(6):1--13.

\bibitem[Uy et~al., 2019]{uy2019revisiting}
Uy, M.~A., Pham, Q.-H., Hua, B.-S., Nguyen, T., and Yeung, S.-K. (2019).
\newblock Revisiting point cloud classification: A new benchmark dataset and
  classification model on real-world data.
\newblock In {\em Proceedings of the IEEE/CVF International Conference on
  Computer Vision}, pages 1588--1597.

\bibitem[Wang et~al., 2017]{wang2017cnn}
Wang, P.-S., Liu, Y., Guo, Y.-X., Sun, C.-Y., and Tong, X. (2017).
\newblock O-cnn: Octree-based convolutional neural networks for 3d shape
  analysis.
\newblock {\em ACM Transactions on Graphics (TOG)}, 36(4):1--11.

\bibitem[Wang et~al., 2018]{wang2018sgpn}
Wang, W., Yu, R., Huang, Q., and Neumann, U. (2018).
\newblock Sgpn: Similarity group proposal network for 3d point cloud instance
  segmentation.
\newblock In {\em Proceedings of the IEEE Conference on Computer Vision and
  Pattern Recognition}, pages 2569--2578.

\bibitem[Wang et~al., 2019a]{wang2019associatively}
Wang, X., Liu, S., Shen, X., Shen, C., and Jia, J. (2019a).
\newblock Associatively segmenting instances and semantics in point clouds.
\newblock In {\em Proceedings of the IEEE Conference on Computer Vision and
  Pattern Recognition}, pages 4096--4105.

\bibitem[Wang et~al., 2019b]{wang2019dynamic}
Wang, Y., Sun, Y., Liu, Z., Sarma, S.~E., Bronstein, M.~M., and Solomon, J.~M.
  (2019b).
\newblock Dynamic graph cnn for learning on point clouds.
\newblock {\em ACM Transactions on Graphics (TOG)}, 38(5):1--12.

\bibitem[Wu et~al., 2019a]{wu2019pq}
Wu, R., Zhuang, Y., Xu, K., Zhang, H., and Chen, B. (2019a).
\newblock Pq-net: A generative part seq2seq network for 3d shapes.
\newblock {\em arXiv preprint arXiv:1911.10949}.

\bibitem[Wu et~al., 2019b]{wu2019sagnet}
Wu, Z., Wang, X., Lin, D., Lischinski, D., Cohen-Or, D., and Huang, H. (2019b).
\newblock Sagnet: Structure-aware generative network for 3d-shape modeling.
\newblock {\em ACM Transactions on Graphics (TOG)}, 38(4):1--14.

\bibitem[Xu et~al., 2017]{Xu_2017_CVPR}
Xu, D., Zhu, Y., Choy, C.~B., and Fei-Fei, L. (2017).
\newblock Scene graph generation by iterative message passing.
\newblock In {\em The IEEE Conference on Computer Vision and Pattern
  Recognition (CVPR)}.

\bibitem[Yang et~al., 2019]{yang2019learning}
Yang, B., Wang, J., Clark, R., Hu, Q., Wang, S., Markham, A., and Trigoni, N.
  (2019).
\newblock Learning object bounding boxes for 3d instance segmentation on point
  clouds.
\newblock In {\em Advances in Neural Information Processing Systems}, pages
  6737--6746.

\bibitem[Yi et~al., 2017]{yi2017learning}
Yi, L., Guibas, L., Hertzmann, A., Kim, V.~G., Su, H., and Yumer, E. (2017).
\newblock Learning hierarchical shape segmentation and labeling from online
  repositories.
\newblock {\em arXiv preprint arXiv:1705.01661}.

\bibitem[Yi et~al., 2016]{yi2016scalable}
Yi, L., Kim, V.~G., Ceylan, D., Shen, I.-C., Yan, M., Su, H., Lu, C., Huang,
  Q., Sheffer, A., and Guibas, L. (2016).
\newblock A scalable active framework for region annotation in 3d shape
  collections.
\newblock {\em ACM Transactions on Graphics (TOG)}, 35(6):1--12.

\bibitem[Yi et~al., 2019]{yi2019gspn}
Yi, L., Zhao, W., Wang, H., Sung, M., and Guibas, L.~J. (2019).
\newblock Gspn: Generative shape proposal network for 3d instance segmentation
  in point cloud.
\newblock In {\em Proceedings of the IEEE Conference on Computer Vision and
  Pattern Recognition}, pages 3947--3956.

\bibitem[Yildirim et~al., 2015]{yildirim2015efficient}
Yildirim, I., Kulkarni, T.~D., Freiwald, W.~A., and Tenenbaum, J.~B. (2015).
\newblock Efficient and robust analysis-by-synthesis in vision: A computational
  framework, behavioral tests, and modeling neuronal representations.
\newblock In {\em Annual conference of the cognitive science society},
  volume~1.

\bibitem[Yuille and Kersten, 2006]{yuille2006vision}
Yuille, A. and Kersten, D. (2006).
\newblock Vision as bayesian inference: analysis by synthesis?
\newblock {\em Trends in cognitive sciences}, 10(7):301--308.

\bibitem[Zacharov et~al., 2019]{zacharov2019zhores}
Zacharov, I., Arslanov, R., Gunin, M., Stefonishin, D., Bykov, A., Pavlov, S.,
  Panarin, O., Maliutin, A., Rykovanov, S., and Fedorov, M. (2019).
\newblock “zhores”—petaflops supercomputer for data-driven modeling,
  machine learning and artificial intelligence installed in skolkovo institute
  of science and technology.
\newblock {\em Open Engineering}, 9(1):512--520.

\bibitem[Zhang and Wonka, 2019]{zhang2019point}
Zhang, B. and Wonka, P. (2019).
\newblock Point cloud instance segmentation using probabilistic embeddings.
\newblock {\em arXiv preprint arXiv:1912.00145}.

\bibitem[Zhang, 2012]{zhang2012microsoft}
Zhang, Z. (2012).
\newblock Microsoft kinect sensor and its effect.
\newblock {\em IEEE multimedia}, 19(2):4--10.

\bibitem[Zhu et~al., 2019]{zhu2019cosegnet}
Zhu, C., Xu, K., Chaudhuri, S., Yi, L., Guibas, L., and Zhang, H. (2019).
\newblock Cosegnet: Deep co-segmentation of 3d shapes with group consistency
  loss.
\newblock {\em arXiv preprint arXiv:1903.10297}.

\bibitem[Zhu and Mumford, 2006]{zhu2006stochastic}
Zhu, S.-C. and Mumford, D. (2006).
\newblock A stochastic grammar of images.
\newblock {\em Foundations and Trends{\textregistered} in Computer Graphics and
  Vision}, 2(4):259--362.

\end{thebibliography}
}

\clearpage

\appendix

\section{Model implementation details}

\paragraph{Network and training}
\label{methods:network_training}
Our models for semantic, instance and hierarchical segmentation tasks are 3D CNNs identical in architecture up to the last layer, which computes voxel features. For each occupied input voxel in a scene, our networks produce a 32-dimensional feature vector. In case of semantic and hierarchical segmentaiton that feature vector is further transformed by a "head" of the model to accommodate a required number of predicted classes. All our networks are fully convolutional, enabling inference for scenes with arbitrary spatial sizes, as long as we have enough RAM and GPU RAM.

\paragraph{discriminative loss}

For \emph{instance segmentation,} the goal is to simultaneously perform part-level semantic labeling and assign each voxel~$v_j$ a unique part instance ID (e.g., to differentiate separate legs of a table).
To this end, we employ a discriminative loss function which has demonstrated its effectiveness in previous works~\cite{de2017semantic,pham2019jsis3d}.

Suppose that there are $K$ instances and $N_k, k\in\{1,...,K\}$ is the number of voxels corresponding to $k$-th instance, $\mathbf{e}_j \in \mathbb{R}^d$ is the embedding of voxel $v_j$, and $\boldsymbol\mu_k$ is the mean of embeddings for the $k$-th instance. The embedding loss $\mathcal{L}_{embedding}$ of a discriminative function can be defined as follows,

\begin{align}
  \label{eq:discriminative}
  \mathcal{L}_{embedding} = \alpha \cdot \mathcal{L}_{pull} + \beta \cdot \mathcal{L}_{push} + \gamma \cdot \mathcal{L}_{reg}
\end{align}
where
\begin{align}
  \label{eq:pull}
  \mathcal{L}_{pull} = \frac{1}{K} \sum_{k=1}^K \frac{1}{N_k} \sum_{j=1}^{N_k} \left [ \left \Vert \boldsymbol\mu_k - \mathbf{e}_j \right \Vert_2 - \delta_v \right ]^2_+
\end{align}
\begin{align}
  \label{eq:push}
  \mathcal{L}_{push} = \frac{1}{K(K-1)} \sum_{k=1}^K \sum_{m=1, m \neq k}^K \left [2\delta_d - \left \Vert \boldsymbol\mu_k - \boldsymbol\mu_m \right \Vert_2 \right ]^2_+
\end{align}
\begin{align}
  \label{eq:reg}
  \mathcal{L}_{reg} = \frac{1}{K} \sum_{k=1}^K \left \Vert \boldsymbol\mu_k \right \Vert_2
\end{align}
where $[x]_+=\max(0,x)$, $\delta_v$ and $\delta_d$ are respectively the margins for the pull loss $\mathcal{L}_{pull}$ and push loss $\mathcal{L}_{push}$. We set $\alpha = \beta = 1$ and $\gamma = 0.001$ the same way as in implementation we used as a reference~\cite{pham2019jsis3d}.

Embedding loss can be understood as follows: the pull loss $\mathcal{L}_{pull}$ pulls embeddings of voxels of some instance towards the centroids, $\boldsymbol\mu_k$, at the same time the push loss $\mathcal{L}_{push}$ pushes these centroids away from each other. The regularisation loss $\mathcal{L}_{reg}$  forces all centroids towards the origin. If we set the margin $\delta_d > 2\delta_v$, very useful property of this loss as described in paper~\cite{de2017semantic} will emerge, the ability to learn the voxel embeddings that will be closer to the instance centroid they belong to, rather than other centroids of other instances.

\section{Dataset details}

You can see detailed stats for the whole dataset at level $d_1$ in Table~\ref{tab:datasetstats}, as you can see it's highly unbalanced. Many semanticly signifficant classes were not included in the dataset because Scan2Cad~\cite{avetisyan2019scan2cad} labels did not include those classes. For more details you can see Table~\ref{tab:scan2part_proportions}.

\begin{table}[!htb]
\centering
\resizebox{0.45\textwidth}{!}{%
\begin{tabular}{l|lllll}
classes & \% in testset & \# voxel & \% voxels & \# scenes with inst. & \# inst. in testset\\ \hline
Microwave & 18.75\% & 20141 & 0.38\% & 90 & 21\\
Display & 19.36\% & 149988 & 2.84\% & 350 & 142\\
Lamp & 20.14\% & 15736 & 0.30\% & 95 & 26\\
Laptop & 19.95\% & 3950 & 0.07\% & 46 & 11\\
Bag & 19.75\% & 25225 & 0.48\% & 136 & 33\\
Storage & 19.41\% & 1833826 & 34.68\% & 866 & 496\\
Bed & 20.94\% & 504774 & 9.55\% & 271 & 69\\
Table & 20.82\% & 1268401 & 23.99\% & 1104 & 539\\
Chair & 16.95\% & 1261928 & 23.87\% & 960 & 754\\
Dishwasher & 20.99\% & 12846 & 0.24\% & 24 & 6\\
TrashCan & 18.96\% & 178648 & 3.38\% & 634 & 199\\
Vase & 20.48\% & 9989 & 0.19\% & 30 & 9\\
Keyboard & 19.63\% & 1819 & 0.03\% & 24 & 9\\
\end{tabular}%
}
\caption{Full dataset and testset statistics.}
\label{tab:datasetstats}
\end{table}

\begin{table}[ht!]
\centering
\resizebox{0.45\textwidth}{!}{%
\begin{tabular}{l|l|l}
Class ID & overlap? & class name \\
\hline
04379243 & 99.0\% (822/830) & \, table \\ \hline
02747177 & 98.9\% (88/89) & \begin{tabular}{l} trash can \\ garbage can \\ wastebin \end{tabular} \\ \hline
03211117 & 97.6\% (161/165) & \begin{tabular}{l} display \\ video display \end{tabular} \\ \hline
03761084 & 97.3\% (36/37) & \begin{tabular}{l} microwave \\ microwave oven \end{tabular} \\ \hline
03337140 & 97.1\% (68/70) & \begin{tabular}{l} file \\ file cabinet \\ filing cabinet \end{tabular} \\ \hline
03001627 & 96.9\% (632/652) & \, chair \\ \hline
02871439 & 96.7\% (145/150) & \, bookshelf \\ \hline
02933112 & 94.8\% (294/310) & \, cabinet \\ \hline
02818832 & 94.0\% (47/50) & \, bed \\ \hline
03991062 & 91.7\% (11/12) & \begin{tabular}{l} pot \\ flowerpot \end{tabular} \\ \hline
03207941 & 85.7\% (12/14) & \begin{tabular}{l} dishwasher \\ dish washer \\ dishwashing machine  \end{tabular} \\ \hline
03085013 & 81.8\% (9/11) & \begin{tabular}{l} computer keyboard \\  keypad \end{tabular} \\ \hline
03325088 & 57.1\% (4/7) & \begin{tabular}{l} faucet \\ spigot \end{tabular} \\ \hline
02876657 & 50.0\% (1/2) & \, bottle \\ \hline
02808440 & 26.0\% (25/96) & \begin{tabular}{l} bathtub \\ bathing tub \\ bath \\ tub \end{tabular} \\ \hline
02801938 & 9.4\% (3/32) & \begin{tabular}{l} basket \\ handbasket \end{tabular} \\ \hline
04256520 & 8.1\% (20/247) & \begin{tabular}{l} sofa \\ couch \\ lounge \end{tabular} \\ \hline
02946921 & 0.0\% (0/1) & \begin{tabular}{l} can \\ tin \\ tin can \end{tabular} \\ \hline
03938244 & 0.0\% (0/5) & \, pillow \\ \hline
02828884 & 0.0\% (0/28) & \, bench \\ \hline
04554684 & 0.0\% (0/37) & \begin{tabular}{l} washer \\ automatic washer \\ washing machine \end{tabular} \\ \hline
03928116 & 0.0\% (0/25) & \begin{tabular}{l} piano \\ pianoforte \\ forte-piano \end{tabular} \\ \hline
03790512 & 0.0\% (0/4) & \begin{tabular}{l} motorcycle \\ bike \end{tabular} \\ \hline
03691459 & 0.0\% (0/2) & \begin{tabular}{l} loudspeaker \\ speaker \\ speaker unit \\ loudspeaker system \end{tabular} \\ \hline
03467517 & 0.0\% (0/6) & \, guitar \\ \hline
04330267 & 0.0\% (0/36) & \, stove \\ \hline
04401088 & 0.0\% (0/1) & \begin{tabular}{l} telephone \\ phone \\ telephone set \end{tabular} \\ \hline
04004475 & 0.0\% (0/31) & \begin{tabular}{l} printer \\ printing machine \end{tabular} \\ \hline
Total:    & 81.2\% (2477/3049) &         \\
\end{tabular}}
\caption{Proportions of ShapeNet semantic classes present in Scan2Cad dataset}
\label{tab:scan2part_proportions}
\end{table}


\subsection{Histograms for all levels of detail}

\begin{figure*}[t]
\centering
\includegraphics[width=0.95\textwidth]{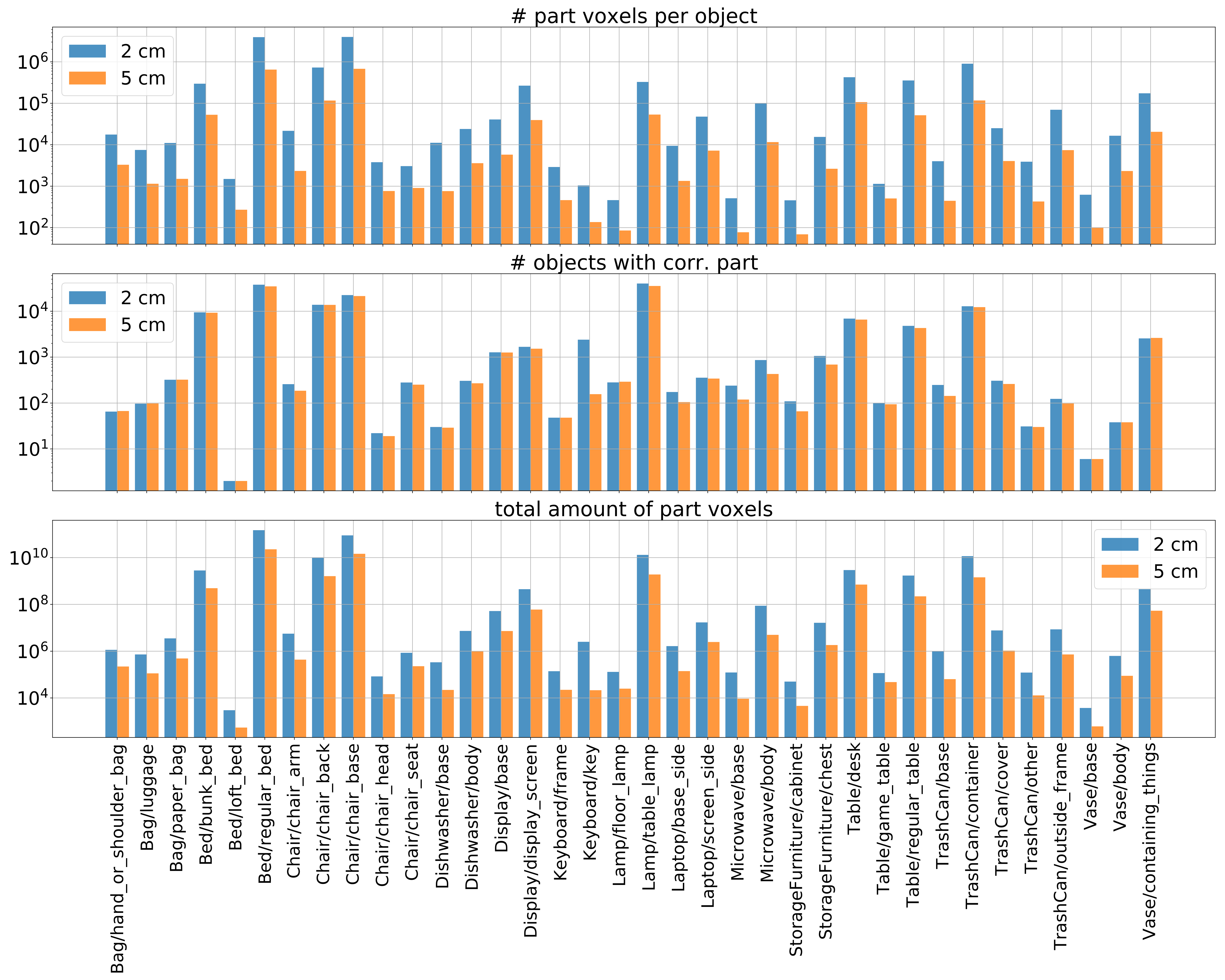}
\caption{Histograms of object statistics for the coarsest ($d = 2$) level of details: TOP: number of voxels per object, CENTER: the number of corresponding objects in Scan2Part, BOTTOM: the total number of object voxels in Scan2Part}
\label{fig:dataset_stata_2}
\end{figure*}

\begin{figure*}[t]
\centering
\includegraphics[width=0.95\textwidth]{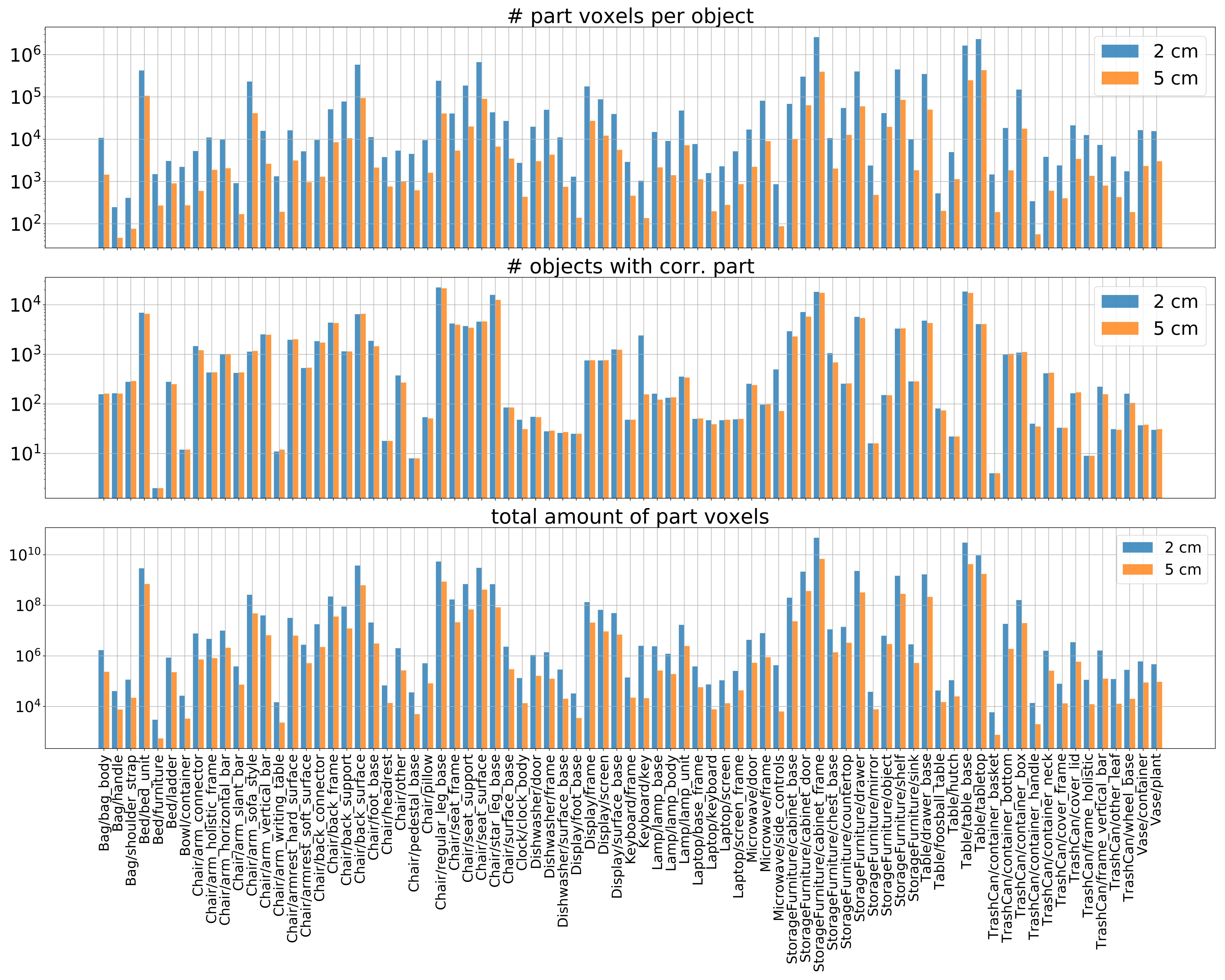}
\caption{Histograms of object statistics for the coarsest ($d = 3$) level of details: TOP: number of voxels per object, CENTER: the number of corresponding objects in Scan2Part, BOTTOM: the total number of object voxels in Scan2Part}
\label{fig:dataset_stata_3}
\end{figure*}

Histograms for levels of detail $d_1$, $d_2$ and $d_3$ can be found respectively in Figure~\ref{fig:dataset_stata_1}, Figure~\ref{fig:dataset_stata_2} and Figure~\ref{fig:dataset_stata_3}.

\begin{figure*}[t]
\centering
\includegraphics[width=0.95\textwidth]{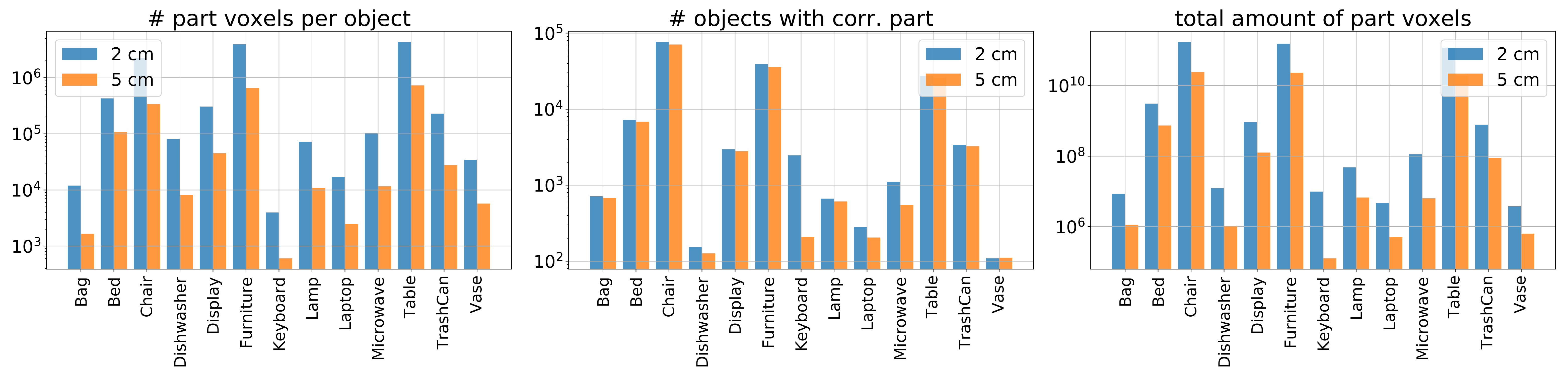}
\caption{Histograms of object statistics for the coarsest ($d = 1$) level of details: LEFT: number of voxels per object, CENTER: the number of corresponding objects in Scan2Part, RIGHT: the total number of object voxels in Scan2Part}
\label{fig:dataset_stata_1}
\end{figure*}

\section{Full results of Segmentation models}

Results for semantic segmentation on level of detail $d_2$ and $d_3$ can be found respectively in Table~\ref{tab:segmentation_for_lod_2} and Table~\ref{tab:segmentation_for_lod_3} .


\begin{table*}[!htp]\centering
\scriptsize
\begin{tabular}{l|rrrrrrrr}\toprule
\textbf{class\_name} &\textbf{3D-Unet} &\textbf{SparseConvNet} &\textbf{Our} &\textbf{Our (w/ color)} &\textbf{MTT (balanced)} &\textbf{MTT (Fine-grained)} &\textbf{MTT (Coarse)} \\\midrule
TrashCan/cover &5.55\% &18.86\% &41.48\% &38.29\% &31.83\% &39.89\% &\textbf{45.80\%} \\
Display/base &11.52\% &36.87\% &\textbf{70.98\%} &51.64\% &68.41\% &54.77\% &57.38\% \\
Table/desk &9.21\% &33.73\% &31.54\% &26.62\% &30.45\% &33.66\% &\textbf{37.17\%} \\
Vase/base &0.00\% &0.00\% &\textbf{0.81\%} &0.00\% &0.00\% &0.00\% &0.00\% \\
Chair/chair\_base &58.00\% &77.53\% &77.70\% &78.89\% &79.29\% &78.97\% &\textbf{81.26\%} \\
Table/regular\_table &49.88\% &71.33\% &73.40\% &71.91\% &74.94\% &68.25\% &\textbf{76.25\%} \\
Bed/loft\_bed &4.71\% &16.70\% &\textbf{16.98\%} &15.37\% &15.50\% &14.87\% &14.51\% \\
Chair/chair\_back &49.96\% &\textbf{72.89\%} &71.89\% &70.79\% &70.11\% &72.03\% &71.91\% \\
Display/display\_screen &41.52\% &70.90\% &70.81\% &70.58\% &\textbf{72.70\%} &69.09\% &70.63\% \\
Bag/paper\_bag &6.97\% &19.93\% &\textbf{41.31\%} &23.93\% &38.51\% &33.85\% &32.40\% \\
TrashCan/outside\_frame &0.00\% &0.00\% &\textbf{2.47\%} &0.24\% &0.00\% &1.49\% &0.00\% \\
Microwave/base &0.00\% &0.00\% &\textbf{4.34\%} &0.10\% &0.09\% &1.32\% &2.04\% \\
Chair/chair\_arm &18.91\% &46.75\% &49.33\% &47.97\% &\textbf{49.77\%} &48.04\% &46.14\% \\
Keyboard/frame &0.13\% &2.92\% &\textbf{5.41\%} &4.12\% &5.10\% &2.90\% &3.66\% \\
StorageFurniture/cabinet &44.10\% &71.24\% &72.12\% &72.39\% &\textbf{72.90\%} &70.52\% &71.14\% \\
TrashCan/container &34.22\% &63.87\% &59.22\% &66.42\% &67.42\% &67.48\% &\textbf{69.36\%} \\
Vase/containing\_things &0.91\% &5.39\% &\textbf{28.99\%} &22.36\% &27.56\% &24.80\% &14.03\% \\
Bed/bunk\_bed &7.20\% &27.46\% &28.48\% &34.53\% &\textbf{37.51\%} &31.32\% &34.07\% \\
StorageFurniture/chest &1.12\% &6.81\% &7.93\% &7.01\% &6.47\% &\textbf{9.70\%} &4.95\% \\
TrashCan/base &7.07\% &27.02\% &19.86\% &18.17\% &25.85\% &\textbf{37.07\%} &30.58\% \\
Bag/hand\_or\_shoulder\_bag &2.56\% &10.35\% &8.91\% &8.04\% &9.64\% &\textbf{23.99\%} &12.83\% \\
Bag/luggage &4.14\% &14.18\% &17.70\% &\textbf{25.14\%} &17.12\% &19.03\% &24.53\% \\
Chair/chair\_seat &50.51\% &74.94\% &77.75\% &75.29\% &\textbf{78.21\%} &74.69\% &75.00\% \\
Lamp/table\_lamp &8.08\% &30.07\% &26.89\% &26.04\% &26.91\% &27.71\% &25.00\% \\
TrashCan/other &0.00\% &0.00\% &5.17\% &3.25\% &\textbf{16.67\%} &0.00\% &0.00\% \\
Laptop/screen\_side &15.63\% &42.76\% &44.94\% &\textbf{54.66\%} &52.93\% &45.85\% &50.53\% \\
Vase/body &0.09\% &2.05\% &1.10\% &1.62\% &4.98\% &7.49\% &\textbf{12.31\%} \\
Bed/regular\_bed &37.46\% &64.43\% &67.19\% &68.07\% &\textbf{72.41\%} &64.36\% &69.13\% \\
Dishwasher/body &0.00\% &0.00\% &0.00\% &0.00\% &\textbf{2.65\%} &0.00\% &0.00\% \\
Chair/chair\_head &0.22\% &4.22\% &5.53\% &3.52\% &\textbf{9.15\%} &7.03\% &8.01\% \\
Table/game\_table &0.00\% &0.00\% &1.47\% &\textbf{12.75\%} &9.43\% &1.47\% &0.00\% \\
Laptop/base\_side &0.00\% &0.00\% &0.00\% &0.00\% &\textbf{0.15\%} &0.00\% &0.00\% \\
Dishwasher/base &0.00\% &0.00\% &0.50\% &0.00\% &\textbf{8.76\%} &0.00\% &1.52\% \\
Microwave/body &1.80\% &8.68\% &14.31\% &20.84\% &\textbf{23.99\%} &11.97\% &8.79\% \\
Keyboard/key &0.00\% &0.00\% &\textbf{2.26\%} &1.52\% &0.00\% &0.00\% &0.00\% \\
Lamp/floor\_lamp &0.00\% &0.00\% &0.00\% &3.17\% &10.00\% &10.87\% &\textbf{15.63\%} \\\midrule
\textbf{Avg.} &13.10\% &25.61\% &29.13\% &28.48\% &31.04\% &29.29\% &29.63\% \\
\bottomrule
\end{tabular}
\caption{IoU for level of detail $d_2$}\label{tab:segmentation_for_lod_2}
\end{table*}

\begin{table*}[!htp]\centering
\scriptsize
\begin{tabular}{l|rrrrrrr}\toprule
\textbf{class\_name} &\textbf{3D-Unet} &\textbf{SparseConvNet} &\textbf{Our (w/ color)} &\textbf{Our} &\textbf{MTT (Fine-grained)} &\textbf{MTT (Coarse)} \\\midrule
Table/hutch &0.00\% &26.11\% &43.76\% &\textbf{49.05\%} &0.00\% &0.00\% \\
StorageFurniture/shelf &19.31\% &39.38\% &\textbf{60.41\%} &54.62\% &53.48\% &56.40\% \\
Bag/shoulder\_strap &1.69\% &4.10\% &7.38\% &10.94\% &21.33\% &\textbf{21.88\%} \\
Display/surface\_base &15.18\% &31.78\% &34.01\% &\textbf{36.57\%} &33.01\% &27.57\% \\
Microwave/frame &0.39\% &0.00\% &\textbf{60.29\%} &56.25\% &0.00\% &0.00\% \\
Chair/seat\_surface &22.01\% &64.09\% &69.63\% &66.43\% &68.36\% &\textbf{75.45\%} \\
StorageFurniture/cabinet\_door &12.45\% &48.38\% &\textbf{50.59\%} &50.27\% &49.57\% &37.14\% \\
TrashCan/container\_box &1.88\% &39.86\% &\textbf{48.82\%} &46.97\% &44.30\% &29.52\% \\
Vase/plant &4.24\% &13.77\% &13.65\% &\textbf{16.74\%} &14.71\% &14.70\% \\
Chair/arm\_horizontal\_bar &10.21\% &15.43\% &18.48\% &\textbf{19.92\%} &17.75\% &14.32\% \\
StorageFurniture/countertop &21.58\% &40.95\% &43.42\% &43.12\% &\textbf{44.15\%} &42.31\% \\
Chair/regular\_leg\_base &26.88\% &57.32\% &61.74\% &\textbf{63.58\%} &61.07\% &56.05\% \\
Chair/seat\_frame &3.94\% &20.14\% &\textbf{33.84\%} &32.96\% &31.26\% &30.65\% \\
Chair/headrest &1.18\% &1.31\% &\textbf{6.87\%} &6.26\% &4.20\% &2.93\% \\
StorageFurniture/cabinet\_base &11.29\% &29.16\% &29.90\% &\textbf{34.69\%} &30.04\% &25.29\% \\
Chair/pillow &0.90\% &2.78\% &\textbf{4.34\%} &2.84\% &3.07\% &3.81\% \\
Bed/bed\_unit &31.86\% &57.43\% &58.91\% &58.75\% &\textbf{59.57\%} &53.09\% \\
Chair/arm\_vertical\_bar &9.86\% &17.99\% &23.16\% &22.93\% &\textbf{23.67\%} &18.55\% \\
Bag/handle &0.54\% &0.52\% &2.77\% &\textbf{9.83\%} &8.96\% &3.58\% \\
Keyboard/key &0.09\% &0.31\% &4.19\% &\textbf{4.47\%} &2.07\% &0.57\% \\
Chair/armrest\_hard\_surface &3.28\% &18.18\% &22.23\% &\textbf{22.32\%} &21.90\% &12.66\% \\
Chair/back\_surface &21.25\% &45.40\% &52.51\% &52.36\% &\textbf{52.69\%} &44.29\% \\
Chair/back\_frame &3.74\% &25.15\% &21.37\% &21.78\% &\textbf{36.44\%} &17.70\% \\
Table/tabletop &46.26\% &73.38\% &72.70\% &73.52\% &74.04\% &\textbf{75.21\%} \\
Bag/bag\_body &2.53\% &19.00\% &24.77\% &\textbf{27.14\%} &22.19\% &13.25\% \\
Clock/clock\_body &2.88\% &4.95\% &7.38\% &8.27\% &\textbf{8.31\%} &5.00\% \\
Laptop/base\_frame &0.00\% &0.00\% &0.00\% &0.32\% &0.34\% &\textbf{2.28\%} \\
TrashCan/container\_neck &0.00\% &30.52\% &31.54\% &34.80\% &\textbf{36.09\%} &26.93\% \\
Chair/back\_connector &1.56\% &13.30\% &19.31\% &\textbf{28.72\%} &27.68\% &11.47\% \\
Keyboard/frame &0.00\% &1.96\% &3.54\% &6.88\% &\textbf{21.67\%} &9.66\% \\
StorageFurniture/sink &0.49\% &14.12\% &\textbf{22.58\%} &9.00\% &17.42\% &19.31\% \\
Dishwasher/frame &0.00\% &4.17\% &0.00\% &0.00\% &\textbf{15.66\%} &9.09\% \\
Chair/other &0.00\% &1.09\% &0.00\% &2.85\% &4.53\% &\textbf{4.67\%} \\
Chair/arm\_slant\_bar &1.92\% &4.50\% &5.34\% &8.36\% &7.85\% &\textbf{9.23\%} \\
Chair/star\_leg\_base &20.65\% &55.77\% &\textbf{56.86\%} &55.80\% &56.69\% &41.82\% \\
Dishwasher/surface\_base &\textbf{0.00\%} &\textbf{0.00\%} &\textbf{0.00\%} &\textbf{0.00\%} &\textbf{0.00\%} &\textbf{0.00\%} \\
Lamp/lamp\_base &4.56\% &17.52\% &\textbf{23.27\%} &22.66\% &21.97\% &20.75\% \\
Dishwasher/door &0.00\% &0.00\% &0.00\% &\textbf{9.95\%} &0.00\% &8.37\% \\
Chair/arm\_connector &3.58\% &17.61\% &\textbf{21.34\%} &19.52\% &20.84\% &14.95\% \\
Chair/seat\_support &10.29\% &26.48\% &28.81\% &28.04\% &\textbf{29.36\%} &27.74\% \\
Laptop/screen &0.25\% &13.35\% &\textbf{20.17\%} &17.90\% &18.08\% &7.39\% \\
StorageFurniture/drawer &21.30\% &29.96\% &32.33\% &33.08\% &33.25\% &\textbf{34.19\%} \\
TrashCan/cover\_frame &0.00\% &0.00\% &0.00\% &0.75\% &0.00\% &\textbf{1.91\%} \\
Laptop/screen\_frame &6.63\% &\textbf{29.45\%} &28.06\% &28.58\% &26.67\% &20.54\% \\
TrashCan/frame\_holistic &\textbf{0.00\%} &\textbf{0.00\%} &\textbf{0.00\%} &\textbf{0.00\%} &\textbf{0.00\%} &\textbf{0.00\%} \\
Chair/armrest\_soft\_surface &0.15\% &4.85\% &\textbf{9.44\%} &3.40\% &5.98\% &7.37\% \\
Vase/container &0.75\% &6.08\% &4.26\% &5.79\% &7.38\% &\textbf{8.09\%} \\
Chair/foot\_base &6.31\% &15.35\% &\textbf{20.07\%} &19.92\% &17.10\% &16.43\% \\
TrashCan/cover\_lid &3.46\% &11.17\% &16.92\% &\textbf{21.07\%} &16.02\% &15.78\% \\
Bed/furniture &0.86\% &5.64\% &17.88\% &19.01\% &\textbf{19.53\%} &14.88\% \\
StorageFurniture/object &3.30\% &11.15\% &11.64\% &\textbf{11.99\%} &11.50\% &7.83\% \\
Bowl/container &\textbf{0.20\%} &0.00\% &0.00\% &0.00\% &0.00\% &0.00\% \\
StorageFurniture/mirror &0.00\% &\textbf{0.25\%} &0.00\% &0.00\% &0.00\% &0.00\% \\
TrashCan/container\_basket &\textbf{0.59\%} &0.00\% &0.00\% &0.00\% &0.00\% &0.00\% \\
Chair/arm\_sofa\_style &11.74\% &39.14\% &\textbf{43.82\%} &43.63\% &39.91\% &36.05\% \\
Chair/arm\_writing\_table &\textbf{0.00\%} &\textbf{0.00\%} &\textbf{0.00\%} &\textbf{0.00\%} &\textbf{0.00\%} &\textbf{0.00\%} \\
StorageFurniture/chest\_base &2.27\% &7.43\% &1.54\% &4.92\% &6.99\% &\textbf{11.69\%} \\
Laptop/keyboard &0.00\% &0.00\% &0.14\% &0.00\% &0.12\% &\textbf{2.46\%} \\
Bed/ladder &3.67\% &6.13\% &11.85\% &\textbf{13.72\%} &12.71\% &10.52\% \\
Display/screen &21.19\% &52.45\% &\textbf{58.53\%} &58.05\% &56.99\% &51.27\% \\
Display/frame &3.66\% &39.40\% &47.35\% &44.02\% &46.65\% &\textbf{49.07\%} \\
Lamp/lamp\_body &5.59\% &13.29\% &15.82\% &24.61\% &27.10\% &\textbf{28.08\%} \\
TrashCan/frame\_vertical\_bar &0.00\% &0.00\% &0.00\% &\textbf{8.79\%} &0.55\% &1.48\% \\
Chair/arm\_holistic\_frame &0.36\% &6.47\% &12.78\% &\textbf{18.35\%} &11.75\% &9.18\% \\
StorageFurniture/cabinet\_frame &10.48\% &40.05\% &\textbf{49.06\%} &46.62\% &47.05\% &46.38\% \\
Table/foosball\_table &4.01\% &0.00\% &0.33\% &3.69\% &1.25\% &\textbf{5.45\%} \\
Chair/surface\_base &0.31\% &3.32\% &3.21\% &\textbf{4.70\%} &3.80\% &2.08\% \\
Lamp/lamp\_unit &8.75\% &41.95\% &48.34\% &49.41\% &\textbf{50.61\%} &47.76\% \\
TrashCan/container\_handle &\textbf{1.83\%} &0.00\% &0.00\% &0.00\% &0.34\% &0.00\% \\
TrashCan/wheel\_base &1.84\% &8.53\% &15.93\% &19.58\% &19.11\% &\textbf{23.91\%} \\
Microwave/side\_controls &0.36\% &5.78\% &9.22\% &10.01\% &\textbf{10.72\%} &5.73\% \\
TrashCan/other\_leaf &\textbf{0.00\%} &\textbf{0.00\%} &\textbf{0.00\%} &\textbf{0.00\%} &\textbf{0.00\%} &\textbf{0.00\%} \\
Chair/pedestal\_base &0.29\% &\textbf{0.43\%} &0.00\% &0.03\% &0.38\% &0.00\% \\
Display/foot\_base &0.16\% &0.38\% &0.47\% &0.12\% &\textbf{1.18\%} &0.00\% \\
Table/drawer\_base &0.13\% &34.39\% &\textbf{40.40\%} &18.47\% &11.52\% &14.64\% \\
TrashCan/container\_bottom &1.27\% &22.24\% &\textbf{38.10\%} &33.33\% &31.38\% &30.92\% \\
Table/table\_base &11.15\% &50.09\% &56.29\% &55.34\% &\textbf{61.13\%} &58.48\% \\
Chair/back\_support &6.09\% &17.87\% &22.83\% &\textbf{23.91\%} &22.07\% &21.66\% \\
Microwave/door &0.00\% &\textbf{12.50\%} &0.00\% &0.00\% &11.90\% &10.26\% \\\midrule
\textbf{Avg.} &5.79\% &17.89\% &21.85\% &\textbf{22.31\%} &21.23\% &18.86\% \\
\bottomrule
\end{tabular}
\caption{IoU for level of detail $d_3$}\label{tab:segmentation_for_lod_3}
\end{table*}

\end{document}